\documentclass{article} % For LaTeX2e
\usepackage{iclr2023_conference,times}

\usepackage{amsmath,amsfonts,bm}

\def\eqref#1{equation~\ref{#1}}
\def\1{\bm{1}}

\DeclareMathAlphabet{\mathsfit}{\encodingdefault}{\sfdefault}{m}{sl}
\SetMathAlphabet{\mathsfit}{bold}{\encodingdefault}{\sfdefault}{bx}{n}

\newcommand{\R}{\mathbb{R}}

\DeclareMathOperator*{\argmin}{arg\,min}

\usepackage{hyperref}
\usepackage{url}

\usepackage{amsthm}
\usepackage{amssymb}
\usepackage{amsmath, mathtools}
\usepackage{bm}
\usepackage{bbm}
\usepackage{booktabs}
\usepackage{diagbox}
\usepackage{dsfont}
\usepackage{caption}
\usepackage{subcaption}
\usepackage{algorithm}
\usepackage{algorithmic}
\usepackage{float}
\usepackage{makecell}
\usepackage{graphicx,wrapfig,lipsum}
\usepackage{soul}
\usepackage{mathtools}

\usepackage{blindtext}

\hypersetup{
	colorlinks=true,
	linkcolor=blue,
	anchorcolor=blue,
	citecolor=blue}

\newtheorem{prop}{Propostition}

\newtheorem{remark}{Remark}

\definecolor{mygreen}{rgb}{0.1,0.75,0.2}

\newcommand{\nc}{\normalcolor}

\newcommand{\NNew }{N^{\text{new}}}

\newcommand{\mlp}{\textsc{MLPNet}}
\newcommand{\mlps}{\textsc{MLPSmall}}
\newcommand{\mlpl}{\textsc{MLPLarge}}
\newcommand{\resnet}{\textsc{ResNet18}}
\newcommand{\vgg}{\textsc{VGG11}}

\title{Wasserstein Barycenter-based Model Fusion and Linear Mode Connectivity of Neural Networks}

\author{Aditya Kumar Akash \thanks{Now at Google} \textsuperscript{ }\thanks{Equal Contribution} \\
Department of Computer Science\\
University of Wisconsin-Madison\\
\texttt{aakash@wisc.edu} \\
\And
Sixu Li \footnotemark[2] \\
Department of Statistics \\
University of Wisconsin-Madison \\
\texttt{sli739@wisc.edu} \\
\AND
Nicolás García Trillos \\
Department of Statistics \\
University of Wisconsin-Madison \\
\texttt{garciatrillo@wisc.edu}
}

\iclrfinalcopy % Uncomment for camera-ready version, but NOT for submission.
\begin{document}

\maketitle

\begin{abstract}
Based on the concepts of Wasserstein barycenter (WB) and Gromov-Wasserstein barycenter (GWB), we propose a unified mathematical framework for neural network (NN) model fusion and utilize it to reveal new insights about the linear mode connectivity of SGD solutions.
In our framework, the fusion occurs in a layer-wise manner and builds on an interpretation of a node in a network as a function of the layer preceding it. 
The versatility of our mathematical framework allows us to talk about model fusion and linear mode connectivity for a broad class of NNs, including fully connected NN, CNN, ResNet, RNN, and LSTM, in each case exploiting the specific structure of the network architecture. 
We present extensive numerical experiments to: 
1) illustrate the strengths of our approach in relation to other model fusion methodologies and 
2) from a certain perspective, provide new empirical evidence for recent conjectures which say that two local minima found by gradient-based methods end up lying on the same basin of the loss landscape after a proper permutation of weights is applied to one of the models. \footnote{Code is available at: \url{https://github.com/SixuLi/WBFusionAndLMC}}
\end{abstract}

\section{Introduction}
\label{sec: introduction}

The increasing use of edge devices like mobile phones, tablets, and vehicles, along with the sophistication in sensors present in them (e.g. cameras, GPS, and accelerometers), has led to the generation of an enormous amount of data. However, data privacy concerns, communication costs, 
bandwidth limits, 
and time sensitivity prevent the gathering of local data from edge devices into one single centralized location. These obstacles have motivated the design and development of federated learning strategies which are aimed at pooling information from locally trained
neural networks (NNs) with the objective of building strong centralized models without relying on the collection of local data \cite{mcmahan2017communication,kairouz2019advances}. 
Due to these considerations, the problem of NN fusion--i.e. combining 
multiple models which were trained differently into a single model--is 
a fundamental task in federated learning.

A standard fusion method for aggregating models with the same architecture is FedAvg \cite{mcmahan2017communication}, which involves element-wise averaging of the parameters of local models. This is also known as vanilla averaging \cite{singh2019model}. Although easily implementable, vanilla averaging performs poorly when fusing models whose weights do not have a one-to-one correspondence. This happens because even when models are trained on the same dataset it is possible to obtain models that differ only by a permutation of weights \cite{wang2020federated, yurochkin2019bayesian}; this feature is known as \textit{permutation invariance property} of neural networks. Moreover, vanilla averaging is not naturally designed to work when using local models with different architectures (e.g., different widths). In order to address these challenges, \cite{singh2019model} proposed to first find the best alignment between the neurons (weights) of different networks by using optimal transport (OT) \cite{villani2008optimal,santambrogio2015optimal,PeyreCuturi} and then carrying out a vanilla averaging step.
In \cite{liu2022deep}, the authors formulate the model fusion as a graph matching problem, which utilizes the second-order similarity of model weights to align neurons. 
% However, this method only works for fully connected (FC) NNs and CNNs without skip-connection.
Other approaches, like those proposed in \cite{yurochkin2019bayesian,wang2020federated}, interpret nodes of local models as random permutations of latent ``global nodes'' modeled according to a Beta-Bernoulli process prior \cite{thibaux2007hierarchical}. By using ``global nodes'', nodes from different input NNs can be embedded into a common space where comparisons and aggregation are meaningful.
Most works in the literature discussing the fusion problem have mainly focused on the aggregation of fully connected (FC) neural networks and CNNs, but have not, for the most part, explored other kinds of architectures like RNNs and LSTMs. One exception to this general state of the art is the work \cite{wang2020federated}, which considers the fusion of RNNs by ignoring hidden-to-hidden weights during the neurons' matching, thus discarding some useful information in the pre-trained RNNs.
For more references on the fusion problem see in the Appendix \ref{sec:Otherfusion}.

{A different line of research that has attracted considerable attention in the past few years is the quest for a comprehensive understanding of the loss landscape of deep neural networks, a fundamental component in studying the optimization and generalization properties of NNs \cite{li2018visualizing,mei2018mean,neyshabur2017exploring,nguyen2018loss,izmailov2018averaging}. Due to over-parameterization, scale, and permutation invariance properties of neural networks, the loss landscapes of DNNs have many local minima \cite{keskar2016large,zhang2021understanding}. Different works have asked and answered affirmatively the question of whether there exist paths of small-increasing loss connecting different local minima found by SGD \cite{garipov2018loss, draxler2018essentially}. This phenomenon is often referred to as mode connectivity \cite{garipov2018loss} and the loss increase along paths between two models is often referred to as (energy) barrier \cite{draxler2018essentially}. It has been observed that low-barrier paths are non-linear, i.e., {linear interpolation of two different models will not usually produce a neural network with small loss}. These observations suggest that, from the perspective of local structure properties of loss landscapes, different SGD solutions belong to different (well-separated) basins \cite{neyshabur2020being}. {However, recent work \cite{entezari2021role} has conjectured that local minima found by SGD do end up lying on the same basin of the loss landscape \textit{after} a proper permutation of weights is applied to one of the models}. 
% With exception of a few works, 
% {\red To the best of our knowledge, this conjecture has not been supported by extensive experimentation in a large variety of settings (architectures and datasets), and 
The question of how to find these desired permutations remains in general elusive.}

\begin{figure}[!htb]
    \centering
    \begin{subfigure}{.4\textwidth}%
    \includegraphics[width=1.0\linewidth]{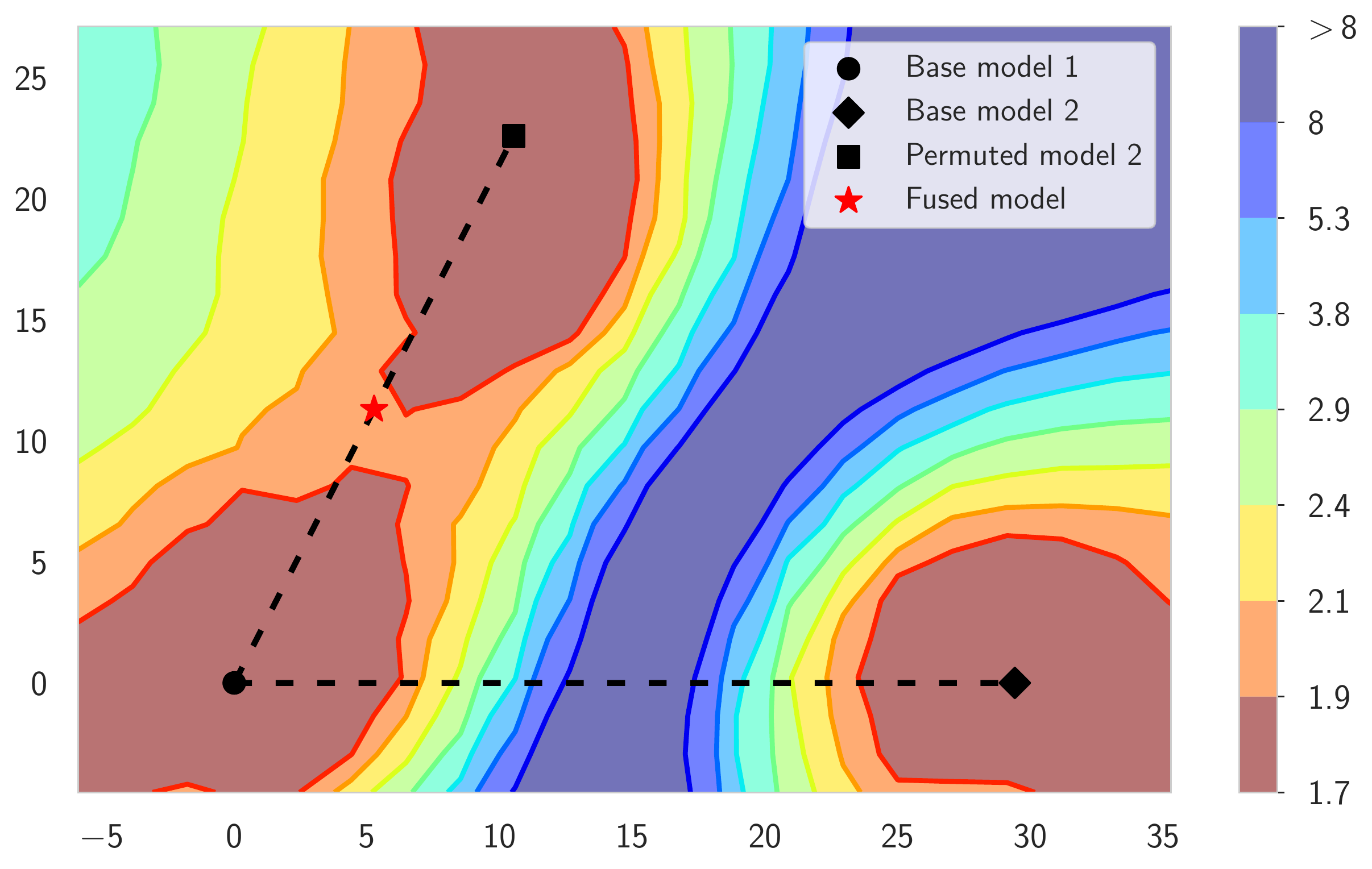}
    \end{subfigure}%
    \begin{subfigure}{.4\textwidth}
    \includegraphics[width=1.0\linewidth]{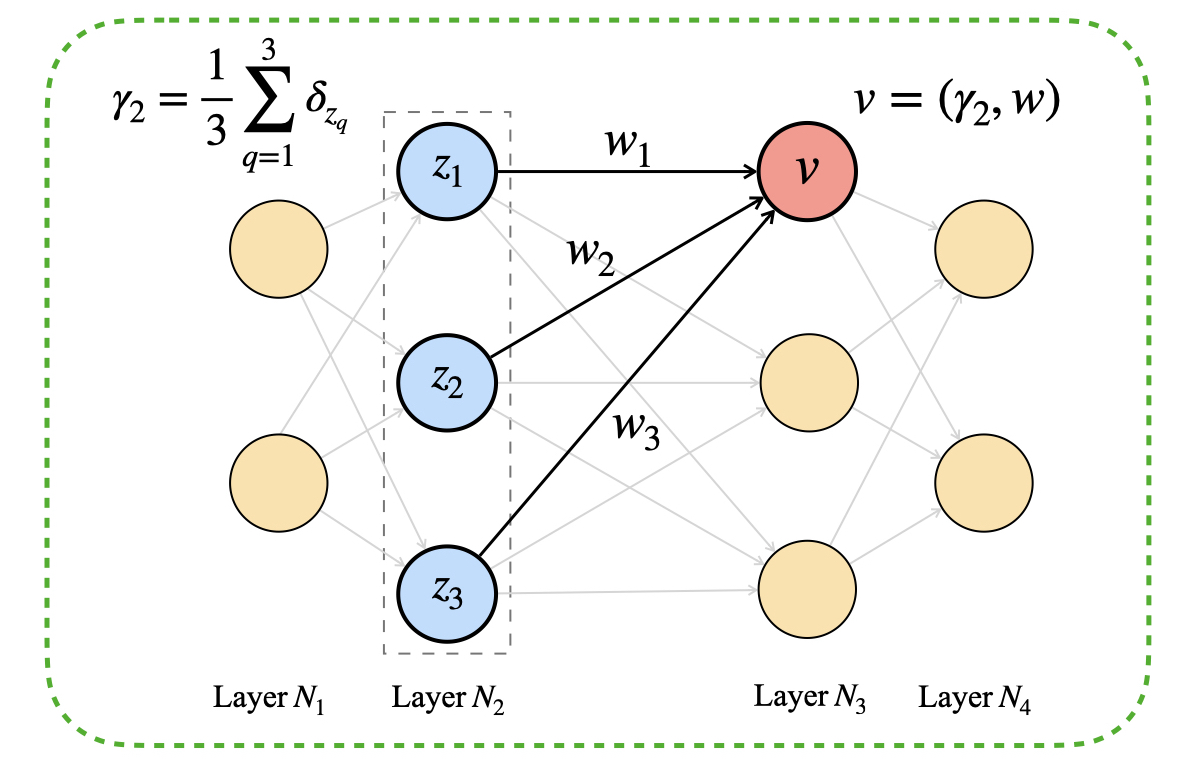}
    \end{subfigure}
\caption{\textbf{Left:} The test error surface of FC NNs trained on MNIST. The permuted model 2 is model 2 \textit{after} multiplying its weights by the coupling obtained by our fusion algorithm. \textbf{Right:} The illustration of our interpretations of FC NNs. Following our definitions, node $v := (\gamma_2, w)$, where $\gamma_2$ is a probability measure on layer $N_2$ and $w: N_2 \rightarrow \mathbb{R}$ is the weight function corresponding to node $v$. For example, the scalar $w(z_{2})$ is the weight between nodes $v$ and $z_{2}$, and we use $w_2$ as the shorthand notation of $w(z_{2})$.}
\label{fig: visualization of FCNN training on MNIST and illustration of FCNN}
\end{figure}

The purpose of this paper is twofold. On one hand, we present a large family of barycenter-based fusion algorithms that can be used to aggregate models within the families of fully connected NNs, CNNs, ResNets, RNNs and LSTMs. The most general family of fusion algorithms that we introduce relies on the concept of Gromov-Wasserstein barycenter (GWB), which allows us to use the information in hidden-to-hidden layers in RNNs and LSTMs in contrast to previous approaches in the literature like that proposed in \cite{wang2020federated}. In order to motivate the GWB based fusion algorithm for RNNs and LSTMs, we first discuss a Wasserstein barycenter (WB) based fusion algorithm for  fully connected, CNN, and ResNet models which follows closely the OT fusion algorithm from \cite{singh2019model}. 
By creating a link between the NN model fusion problem and the problem of computing Wasserstein (or Gromov-Wasserstein) barycenters, our aim is to exploit the many tools that have been developed in the last decade for the computation of WB (or GWB) —see the Appendix \ref{sec: computational OT} for references— and to leverage the mathematical structure of OT problems. 
Using our framework, we are able to fuse models with different architectures and build target models with arbitrary specified dimensions (at least in terms of width). 
On the other hand, through several numerical experiments in a variety of settings (architectures and datasets), we provide new evidence backing certain aspects of the conjecture put forward in \cite{entezari2021role} about the local structure of NNs' loss landscapes. {Indeed, we find out that there exist sparse couplings between different models that can map different local minima found by SGD into basins that are only separated by low energy barriers.} 
These sparse couplings, which can be thought of as approximations to actual permutations, are obtained using our fusion algorithms, which, surprisingly, only use {training data} to set the values of some hyperparameters. We explore this conjecture in imaging and natural language processing (NLP) tasks and provide visualizations of our findings.
{Consider, for example, Figure \ref{fig: visualization of FCNN training on MNIST and illustration of FCNN} (left), which is the visualization of fusing two FC NNs independently trained on the MNIST dataset. We can observe that the basins where model 1 and permuted model 2 (i.e. model 2 \textit{after} multiplying its weights by the coupling obtained by our fusion algorithm) land are close to each other and are only separated by low energy barriers.}

{Our \textbf{main contributions} can then be summarized as follows: 
% \textbf{(a)} we introduce a  mathematical framework to describe NNs using ideas from OT;
\textbf{(a)} we formulate the network model fusion problem as a series of Wasserstein (Gromov-Wasserstein) barycenter problems, bridging in this way the NN fusion problem with computational OT; {\textbf{(b)} we empirically demonstrate that our framework is highly effective at fusing different types of networks, including RNNs and LSTMs.} \textbf{(c)} we visualize the result of our fusion algorithm when aggregating two neural networks in a 2D-plane. By doing this we not only provide some illustrations on how our fusion algorithms perform, but also present empirical evidence for the conjecture made in \cite{entezari2021role}, casting light over the loss landscape of a variety of neural networks. }

{At the time of completing this work, we became aware of two very recent preprints which also explore the conjecture made in \cite{entezari2021role} empirically. In particular, \cite{ainsworth2022git} demonstrates that there is zero-barrier LMC (after permutation) between two independently trained NNs (including ResNet) provided the width of layers is large enough. In \cite{benzing2022random}, the conjecture is explored for FC NNs, finding that the average of two randomly initialized models using the permutation revealed through training gives a non-trivial NN. Compared to our work, none of these two works explored this conjecture for recurrent NNs; we highlight that our GWB fusion method is of particular relevance for this aim. To the best of our knowledge, we thus provide the first-ever exploration of the conjecture posited in \cite{entezari2021role} for NLP tasks.}

\subsection{Notation}
{We first introduce some basic notation and briefly review a few relevant concepts from OT. 
A simplex of histograms with $n$ bins is denoted 
by $\Sigma_n := \{a \in \mathbb{R}_+^n: \sum_i a_i =1\}$.}
%The entropy of $\Pi:= \big[ \pi_{ij}\big]_{i,j} \in \mathbb{R}_+^{n \times n}$ is defined as $H(\Pi) := -\sum_{i,j=1}^{n} \pi_{ij}(\log \pi_{ij} - 1)$. 
The set of couplings between histograms $a \in \Sigma_{n_1}$ and $b \in \Sigma_{n_2}$ 
is denoted by
$\Gamma(a,b) := \{\Pi \in \mathbb{R}_+^{n_1 \times n_2}: \Pi\mathds{1}_{n_2} = a, \Pi^T\mathds{1}_{n_1} = b\}$, where $\mathds{1}_{n} := (1, \dots, 1)^T \in \mathbb{R}^n$. For any 4-way tensor $\mathcal{L} = \big[\mathcal{L}_{ijkl}\big]_{i,j,k,l}$ and matrix $\Pi = \big[\pi_{ij}\big]_{i,j}$, we define the tensor-matrix multiplication of $\mathcal{L}$ and $\Pi$ as the matrix $\mathcal{L} \otimes \Pi := \big[\sum_{k,l} \mathcal{L}_{ijkl}\pi_{kl}\big]_{i,j}$. 

\subsection{Optimal transport and Wasserstein barycenters}\label{sec:WassBarycenters}
Let $\mathcal{X}$ be an arbitrary topological space and let $c: \mathcal{X} \times \mathcal{X} \rightarrow [0,\infty)$ be a cost function assumed to satisfy $c(x,x)=0$ for every $x$. We denote by $\mathcal{M}_1^+(\mathcal{X})$ the space of (Borel)  probability measures on $\mathcal{X}$. For $\{x_i\}_{i=1}^{n_1}, \{y_j\}_{j=1}^{n_2} \in \mathcal{X}$, define discrete measures $\mu = \sum_{i=1}^{n_1}a_i \delta_{x_i}$ and $\nu = \sum_{j=1}^{n_2} b_j \delta_{y_j}$ in $\mathcal{M}_1^+(\mathcal{X})$, where $a \in \Sigma_{n_1}$, $b \in \Sigma_{n_2}$, and $\delta_x$ denotes the Dirac delta measure at $x \in \mathcal{X}$. The Wasserstein ``distance'' between $\mu$ and $\nu$, relative to the cost $c$, is defined as
\begin{equation}\label{Wasserstein distance}
		W(\mu, \nu) := \inf_{\Pi \in \Gamma(\mu, \nu)} \langle C, \Pi\rangle,
\end{equation}
where $C:= \big[c(x_i, y_j)\big]_{i,j}$ is the ``cost'' matrix between $\{x_i\}_i$, $\{y_j\}_j \in \mathcal{X}$, $\Pi := \big[\pi_{ij}\big]_{i,j}  \in \Gamma(\mu, \nu)$
% \footnote{In the sequel we will abuse the notation slightly and use $\Gamma(a, b)$ and $\Gamma(\mu, \nu)$ interchangeably.} 
is the coupling matrix between $\mu$ and $\nu$, and $\langle A, B\rangle := \text{tr}(A^TB)$ is the Frobenius inner product.
	
Let $\{\gamma^i\}_{i=1}^{n} \in \mathcal{M}_1^+(\mathcal{X})$ be a collection of discrete probability measures. The Wasserstein barycenter problem (WBP) \cite{agueh2011barycenters} associated with these measures reads
	\begin{equation}\label{eq: standard WB}
		\min_{\gamma \in \mathcal{M}_1^+(\mathcal{X})} \frac{1}{n}\sum_{i=1}^{n} W(\gamma, \gamma^i).
	\end{equation}
A minimizer of this problem is called a Wasserstein barycenter (WB) of the measures $\{\gamma^i\}_{i=1}^{n}$ and can be understood as an average of the input measures. 
{In the sequel we will use the concept of WB to define fusion algorithms for FC NN, CNN, and ResNet. 
% See Section \ref{main problem} for the FC case and the Supplement for extensions to CNN and ResNet. 
{For RNN and LSTM the fusion reduces to
%When extending the idea of WB to the fusion of RNN and LSTM, it  eventually boils down to 
solving a series of Gromov-Wasserstein barycenter-like problems (see the reviews of GWBP in the Appendix).}}

\section{Wasserstein barycenter based fusion}\label{main problem}
In this section, we discuss our layer-wise fusion algorithm based on the concept of WB. 
First we introduce the necessary interpretations of nodes and layers of NNs in Section \ref{sec: measure on neural networks}. 
Next in Section \ref{sec: cost function}, we describe how to compare layers and nodes across different NNs so as to make sense of aggregating models through WB. Finally we present our fusion algorithm in Section \ref{sec: WB fusion algorithm}.

\subsection{Nested definition of fully connected NN }\label{sec: measure on neural networks}

% In this section, we provide a suitable interpretation for nodes of a NN {that forms the basis of our model fusion algorithm.}
%We interpret node $v_{l,h}^i$ {in the $l$-th layer} (for $l>1$) 
%as an element that couples a function on the domain $N^i_{l-1}$ {(previous layer)} with a probability measure. 
{For a \textit{fully connected} network $N$, we use $v$ to index the nodes in its $l$-th layer $N_l$. 
Let $\gamma_{l}$ denote a probability measure on the $l$-th layer defined as the weighted sum of Dirac delta measure over the nodes in that layer, i.e.,
\begin{equation}\label{def: definition of measure on layers}
	\gamma_{l} :=  \frac{1}{|N_{l}|}\sum_{v \in N_{l}} \delta_{v} \in \mathcal{M}_1^+(N_{l}).
\end{equation}

% For the first layer, we define nodes to index the order of input features. 
We interpret a node $v$ from the $l$-th layer
as an element in $N_l$
that couples a function on the domain $N_{l-1}$ {(previous layer)} 
with a probability measure.}
{In particular, the node $v$ is interpreted as $v := (\gamma_{l-1}, w)$, 
where $\gamma_{l-1}$ is a measure on the previous layer $N_{l-1}$ 
and $w$ represents the weights between the node $v$ and the nodes in previous layer $N_{l-1}$.
{These weights can be interpreted as a function $w: N_{l-1} \rightarrow \R$ and we use the notation $w_q$ to denote the value of function $w$ evaluated at the $q$-th node in the previous layer $N_{l-1}$.}
For the first layer i.e. $l=1$, the nodes simply represent placeholders for the input features.}
The above interpretation is illustrated in Figure \ref{fig: visualization of FCNN training on MNIST and illustration of FCNN} (right).  
This interpretation of associating nodes with a function of previous layer allows us to later define ``distance'' between nodes in different NNs (see Section \ref{sec: cost function}) and 
is motivated from $TL^p$ spaces and distance \cite{GarciaTrillos2015,thorpe2017transportation}
which is designed for comparing signals with different domains (see more details in Appendix \ref{sec:TLpsub}).  
% {\blue This gives an inductive definition of the nodes 
% where the first one is indexing input feature.}
% is particularly useful when comparing 
% the nodes between different NNs. It is helpful especially when the NNs have different 
% architectures.
% WE SHOULD ADDRESS THE SPECIFIC CHOICE OF MEASURE FOR A NODE SO ADDRESS ANY QUESTIONS WE MIGHT HAVE ON IT LATER!!!!
%The first component $\gamma_{l-1}^i$ is a measure on the previous layer $N_{l-1}^i$.} 
%For any layer $l$, the measure $\gamma_l^i$ is defined as weighted sum of Dirac delta measure over the nodes on layer $N_l^i$, i.e.,
%\begin{equation}\label{def: definition of measure on layers}
%	\gamma_{l}^i :=  \frac{1}{k_l^i}\sum_{h=1}^{k_{l}^i} \delta_{v_{l, h}^i} \in \mathcal{M}_1^+(N_{l}^i)
%\end{equation}
% where $b_{l,h}^i$ is a mass value assigned to node $v_{l,h}^i$, and $\sum_{h=1}^{k_l^i}b_{l,h}^i = 1$. For simplicity, in the remainder we restrict our attention to the choice $b_{l,h}^i = \frac{1}{k_l^i}$, but other choices of $b_{l,h}^i$ are possible. 
% {\red Note that the domains of the weight functions $W^i_{l,h}$ are different for nodes from different networks. However, a cost function, inductively defined and introduced later on, induces an OT problem that allows us to compare neurons in different models. }
% Motivation for this specific interpretation of a node can be found in the Supplement.
%{\red \begin{remark}
%	Notice that since the elements of $N_{l}^i$ are interpreted as elements in $TL^2(N_{l-1}^i)$ (for $l>1$),  we have $   \gamma_{l}^i  \in \mathcal{M}_1^+ (TL^2(N_{l-1}^i))$ for all $l>1$. 
%\end{remark}}

\subsection{ Cost functions for comparing layers and nodes}\label{sec: cost function}
% Our construction for nodes and layers of a NN 
% is motivated by the fact that it enables us to 
% compare and define distances between layers of different
% NNs {\blue without assuming anything on the architecture.} 
% This is crucial for model fusion.
% To make sense of an ``average'' $l$-th layer which is 
% equally close to $l$-th layers from different NNs 
% we need to first describe distances between layers.
Having introduced our interpretations of NNs, we now define the cost functions for comparing layers and nodes which will be used to aggregate models through WB.
Consider the $l$-th layers $N_l$ and $N_l'$ of two NNs $N$ and $N'$ respectively. 
We use Wasserstein distance between the measures $\gamma_{l}$ and $\gamma_{l}'$ over $N_l$ and $N_l'$ respectively to define distance between the layers:
\begin{equation}\label{eq: dMu}
		d_{\mu}(\gamma_{l}, \gamma_{l}') := W(\gamma_{l}, \gamma_{l}') = \inf_{\Pi_{l} \in \Gamma(\gamma_{l}, \gamma_{l}')} \langle C_{l}, \Pi_{l}\rangle
\end{equation}
where matrix $\Pi_l = [\pi_{l, jg}]_{j,g}$ is a \textit{coupling} {between the measures $\gamma_l$ and $\gamma_l'$}; and
$C_l$ is the cost matrix give by $C_l := \big[c_l(v, v')\big]_{v, v'}$, where $c_l$ is a cost function between nodes on the $l$-th layers. 

%The above defined distance between layers depends on 
%the choice of cost function 
%between nodes from different NNs.

Following our inductive interpretation of NNs, 
the cost function $c_l$ can also be defined inductively.
Consider nodes $v$ and $v'$ from $l$-th layer of NNs
$N$ and $N'$ respectively.
For the first layer $l=1$, we pick a natural candidate
for cost function, namely $c_1(v, v') : = \mathds{1}_{v \not = v'}$, a reasonable choice given that all  networks have the same input layer. 
{For $l \geq 2$, 
%we can treat nodes $v, v'$ as two signals support on different domains. In particular, 
recall our interpretation of nodes $v = (\gamma_{l-1}, w), v' = (\gamma_{l-1}', w')$, where $\gamma_{l-1}$ and $\gamma'_{l-1}$ denotes the respective
measures associated with previous layer $l-1$ and $w, w'$ denotes the respective weight functions for nodes $v$ and $v'$. 
Since the domains of the weight functions $w$ and $w'$ 
are layers in different NNs, 
it is not clear how to compare them directly. 
However in $TL^p$ interpretation,
after finding a suitable coupling between the support measures $\gamma_{l-1}$ and
$\gamma'_{l-1}$, one can couple the functions $w$ and $w'$ and use a direct L2-comparison.
Motivated by computational and methodological considerations, we use a 
slight modification of the $TL^p$ distance and decouple the problem for the measures from the weights.
Specifically, we define $c_l(v, v') := d_{\mu}(\gamma_{l-1}, \gamma_{l-1}') + d_{W}(w, w')$; where $d_{\mu}$ is the Wasserstein distance (as defined in \eqref{eq: dMu}) between the measures $\gamma_{l-1}$ and $\gamma_{l-1}'$ from layers $l-1$. And $d_W$ is defined using the \textit{optimal coupling} of weight functions' support measures, i.e.,
\begin{equation}\label{eqn: d_W}
    d_W(w, w') := \sum_{q,s} \big(w_q - w'_s \big)^2 (\pi_{l-1,qs})^* =: \langle L(w, w'), (\Pi_{l-1})^* \rangle,
\end{equation}
where 
$L(w, w') := \big[(w_q - w'_s)^2 \big]_{q,s}$ and 
$(\Pi_{l-1})^* = \big[(\pi_{l-1, qs})^* \big]_{q,s}$ is the optimal coupling between $\gamma_{l-1}$ and $\gamma_{l-1}'$. Note that $d_{\mu}(\gamma_{l-1}, \gamma_{l-1}')$ is a fixed constant when comparing any two nodes on the $l$-th layers $N_l$ and $N_l'$. For simplicity, we let $c_l(v, v') = d_W(W,W')$ in what follows, and the information of support measures $\gamma_{l-1}$ and $\gamma_{l-1}'$ is implicitly included in their optimal coupling $(\Pi_{l-1})^*$.}
{Here we have omitted bias terms to ease the exposition of our framework, but a natural implementation that accounts for bias terms can be obtained by simply concatenating them with the weight matrix.}
% Therefore our cost function naturally takes the form $c_l(v, v') := d_{\mu}(\gamma_{l-1}, \gamma_{l-1}') + d_{W}(W, W')$. 
% Here $d_\mu$ is the Wasserstein distance between the measures $\gamma_{l-1}$ and $\gamma'_{l-1}$ from the previous layers and 
% {$d_W$ measures the discrepancy between weights functions.}
% % using the couplings
% % from $d_\mu$.
% This choice of cost function is motivated by the $TL^p$ distance introduced in \cite{GarciaTrillos2015} and further explored in \cite{thorpe2017transportation,LinearTLp}, which takes into account a base measure discrepancy in order to compare signals supported on different domains; see the Supplement. Motivated by computational and methodological considerations, here we use a modification of the $TL^p$ distance and decouple the problem for the measures from the weights. [EXPLAIN BETTER HERE]

% (This sentence needs to roughly change) Let $\mathcal{L}(W_l, W_l^i):= \big[ \big(W_{l,j}(z_{l-1, q}) - W_{l,h}^i(v_{l-1,s}^i)\big)^2\big]_{j,h,q,s}$. Since our definition is inductive and we compare 
% layerwise, the solution $(\Pi^i_{l-1})^*$ to [prev equation] for the previous layer, we now define the cost function $d_{W}$ as:
% \begin{equation}\label{eq: dW}
% 			d_{W}(W, W') := \mathcal{L}(W, W') \otimes (\Pi_{l-1})^*,
% \end{equation}
% Note that this choices makes $d_\mu()$ term a constant
% and for sake of clarify we define $c_l() = d_W()$ in subsequent which $\Pi_{l-1}^{*}$ implicit unless specified.

{We set $(\Pi_{1})^*$ equal to the identity matrix normalized by the size of input layer given that this is a solution to \eqref{eq: dMu} when the cost $c_{1}$ is defined as $c_{1}(v,\tilde v) : = \mathds{1}_{v \not = \tilde v}$. {Other choices of cost function $c_l$ are possible, e.g. the activation-based cost function proposed in \cite{singh2019model}.}}

% \subsection{Fusion at the $l$-th layer}
%We now use the discussion in Section \ref{measure on neural networks} to describe an
%inductive construction of the layers of the target network $\NNew$. 
\subsection{Fusion algorithm}\label{sec: WB fusion algorithm}
{In the following we consider $n$ input FC NNs $N^1, \dots, N^n$.
%that are assumed to have the same number of layers $m$. 
We use $N_l^i$ to denote the $l$-th layer of the $i$-th network $N^i$ and 
$k_l^i$ to denote the number of nodes in that layer, i.e. $ k_l^i=|N_l^i|$. 
Let $\gamma_l^i$ to be the probability measure on layer $N_l^i$ similar to definition in \eqref{def: definition of measure on layers} with the support points being nodes in that layer.}
% For simplicity of notations, we drop subscript $l$ from $v_{l,h}^i$ and denote it using $v_{h}^i$.
% Let $v_{l,h}^i$ to denote the $h$-th node in layer $N_l^i$.
We denote the target model (i.e. the desired fusion output) by $\NNew$ and 
use $k_2, \dots, k_m$ to denote the sizes of its layers $\NNew_2, \dots, \NNew_m $, respectively.
We assume that all networks, including the target model, have the same input layer and the same number of layers $m$.
%The input layer for $\NNew$ is assumed to be the same as that of the input models. 
% For simplicity, throughout the paper we also assume that the bias terms for all networks are set to zero. 
% {\red An adaptation of our algorithm to account for the bias term is straightforward and details are omitted for brevity.}
% A straightforward modification of our set-up allows us to treat the general case.

{Based on the discussion in Sections \ref{sec: measure on neural networks} and \ref{sec: cost function}, we now describe an
inductive construction of the layers of the target network $\NNew$ by fusing all $n$ input NNs.
First, $\NNew _1$ is set to be equal to $N^1_1$: 
this is the base case of the inductive construction 
and simply means that we set the input layer of $\NNew$ to be the same as that of the other models; we also set $\gamma_1:=\gamma_1^1$. 
% For $l>1$ we will eventually interpret a node $z$ in $N_l^{\text{new}}$ as an element in the space $TL^2(N_{l-1}^{\text{new}})$ (notice that in this representation the weights between layer $l-1$ and $z$ are implicitly specified).
Next, assuming that the fusion has been completed for the layers $1$ to $l-1$ ($l\geq 2$), we consider the fusion of the $l$-th layer. For the simplicity of notations, we drop the index $l$ while referring to nodes and their corresponding weights in this layer. 
In particular, we use $v_{g}^i$ and $w_{g}^i$ to denote the nodes in layer $N_l^i$ and their corresponding weights.
To carry out fusion of the $l$-th layer of the input models, 
we aggregate their corresponding measures
through finding WB 
which provides us with a sensible ``average'' $l$-th layer for the target model. 
Hence, we consider the following WBP over $\gamma_l^1, \dots, \gamma_l^n$:
%We consider the following WB optimization problem to carry out the fusion of the $l$-th layers of the input models :
{
	\begin{equation}\label{eq: formal WB on l-th layer}
			\min_{\gamma_l, \{\Pi_l^i\}_i} \quad \frac{1}{n} \sum_{i=1}^{n}W(\gamma_{l}, \gamma_{l}^i) := \frac{1}{n} \sum_{i=1}^{n} \langle C_l^i, \Pi_l^i\rangle
			\qquad \text{s.t.} \,\,\,  \gamma_{l} = \frac{1}{k_l} \sum_{j=1}^{k_l} \delta_{v_{j}}, \,\, v_{j} = (\gamma_{l-1}, w_{j}).
	\end{equation}
% Here each of the matrices $\Pi_l^i = [\pi_{l, jh}^i]_{j,h}$ can be interpreted as a \textit{coupling} {between the measures $\gamma_l$ (the target) and $\gamma_l^i$}. 
% we recall that the measures $\gamma_l^i$ are defined in  \eqref{def: definition of measure on layers}. 
%	 These measures can all be interpreted as measures on $TL^2(\Nold_{l-1})$ (see Remark \ref{rem:MeasuresGamma} below). 
Here the measure $\gamma_l$ is the candidate  $l$-th layer ``average'' of the input models and is forced to take a specific form (notice that we have fixed the size of its support and the masses assigned to its support points).
Nodes $v_{j}$ in the support of $\gamma_l$ 
are set to take the form $v_{j}=(\gamma_{l-1},w_{j})$, i.e. the measure $\gamma_{l-1}$ obtained when fusing the $(l-1)$-th layers is the first coordinate in all the $v_{j}$. 
% {\blue in close correspondence to the structure of the input models.} 
This plugs the current layer of the target model with its previous layer.
As done for the input models, $w_{j}$ is interpreted as a function from the $(l-1)$-th layer into the reals, and represents the actual weight vector from the $(l-1)$-layer to the $j$-th node in the $l$-th layer of the new model. $C_l^i := \big[c_l(v_{j}, v_{g}^i)\big]_{j, g}$ are the cost matrices corresponding to WBP in \eqref{eq: formal WB on l-th layer}, where $c_l$ is a cost function between nodes on the $l$-th layers (see in Section \ref{sec: cost function}). 
Let $W_l$ and $W_l^i$ to be the weight function matrices of the $l$-th layer of target models $N^{\text{new}}$ and input model $N^i$ respectively (e.g. $W_l := (w_{1}, \dots, w_{k_l})^T$) 
and define $\mathcal{L}(W_l, W_l^i):= \big[ \big(w_{jq} - w_{gs}^i \big)^2 \big]_{j, g, q,s}$, where $w_{jq}$ denotes the function $w_{j}$ evaluated at the $q$-th node in layer $l-1$ and similarly for $w_{gs}^i$. The cost matrices $C_l^i$ can now be rewritten as
\begin{equation}\label{eqn: cost matrix}
    C_l^i := \big[c_l(v_{j}, v_{g}^i)\big]_{j, g} = \big[d_W(w_{j}, w_{g}^i) \big]_{j,g} = \mathcal{L}(W_l, W_{l}^i) \otimes (\Pi_{l-1}^i)^*,
\end{equation}
where $(\Pi_{l-1}^i)^*$ is the \textit{optimal coupling} between measures $\gamma_{l-1}$ and $\gamma_{l-1}^i$.}
Combining \eqref{eqn: cost matrix} with \eqref{eq: formal WB on l-th layer} gives us the following optimization problem which we solve to obtain the fused layer:
\begin{equation}\label{eq: problem after plugging in cost function d_W}
		\min_{W_l, \{\Pi_l^i\}_i} B(W_l; \{\Pi_l^i\}_i) := \frac{1}{n} \sum_{i=1}^{n}\langle \mathcal{L}(W_l, W_l^i) \otimes (\Pi_{l-1}^i)^*, \Pi_{l}^i\rangle.
\end{equation}}
In order to solve the minimization problem \ref{eq: problem after plugging in cost function d_W}, we can follow a strategy discussed in \cite{cuturi2014fast,anderes2016discrete,claici2018stochastic}, i.e., alternatingly update weights $W_l$ and couplings $\{\Pi_l^i\}_i$ (remember that the $(\Pi_{l-1}^i)^*$ are computed once and for all and are fixed in \eqref{eq: problem after plugging in cost function d_W}. In particular, after initializing weight matrices, we alternate between two steps until some stopping criterion is reached:

\textbf{Step 1}:
For fixed $W_l$, we update the couplings $\{\Pi_l^i\}_i$. Note that the minimization of $B(W_l; \{\Pi_l^i\}_i)$  over the couplings $\{\Pi_l^i\}_i$ splits into $n$ OT problems, each of which can be solved using any of the algorithms used in computational OT (e.g. Sinkhorn's algorithm \cite{cuturi2013sinkhorn}). 
% see a review in the Supplement). 

\textbf{Step 2}: For fixed couplings $\{\Pi_l^i\}_i$, we update the weights $W_l$. {Note} that for fixed couplings the objective $B(W_l; \{\Pi_l^i\}_i)$ is quadratic in $W_l$ 
{and hence we obtain the following update formula:} 
\begin{equation}\label{eq: update formula for W_l}
		W_l \leftarrow k_l k_{l-1} \frac{1}{\mathds{1}_{k_{l-1}}\mathds{1}_{k_l}^T} \frac{1}{n} \sum_{i=1}^{n} \Pi_{l}^i  W_l^i (\Pi_{l-1}^i)^{*T},
\end{equation}
where $\frac{\cdot}{\cdot}$ is elementwise division. We refer to the above fusion algorithm as \textit{Wasserstein barycenter-based fusion} (WB fusion). 
{The pseudo-code for this algorithm and corresponding computational complexity can be found in the Appendix \ref{sec: detailed WB fusion}, where we also provide some details on how to adapt our fusion method to handle convolutional layers and skip-connections.}

\section{Gromov-Wasserstein barycenter-based fusion}\label{sec:GWFusion}

% \begin{wrapfigure}{R}{0.5\textwidth}
%     \vspace{-15pt}
% 	\centering
% 	\includegraphics[width=0.5\textwidth]{pics/rnn.png}
% 	\caption{A building block of unfolded RNN motivating problem \eqref{eq: GWBP for rnn fusion}. $W_l^{(t)}$ and $H_l^{(t)}$ are collections of input-to-hidden and hidden-to-hidden weights functions at time step $t$ respectively (for actual RNNs, $W_l$ and $H_l$ don't depend on $t$). $x_{l}^{(t)}$ denotes the output of layer $N_l$ at time step $t$; $t$, which changes horizontally, indexes unfolded units.}
% 	\label{fig: RNN's building block}
% 	\vspace{-12pt}
% \end{wrapfigure}
 In this section we discuss extension of our fusion framework to cover RNNs and LSTMs. Compared to FC networks, RNNs contain ``self-loops'' in each layer (hidden-to-hidden recurrent connections) which allows information to be passed from one step of the neural network to the next. 
 Similar to our interpretation of neurons in the FC case, a node $v_g^i$ on the $l$-th layer will be represented as $v_g^i := \big[(\gamma_{l-1}^i, w_{g}^i); (\gamma_{l}^i, h_{g}^i)\big]$, where $w_{g}^i$ is the weight function between inputs of the preceding layer and hidden states, and $h_{g}^i$ is the weight function between hidden states; $\gamma_{l-1}^i$ and $\gamma_{l}^i$ are the probability measures corresponding to layer $l-1$ and layer $l$ respectively. This definition comes from the observation that hidden-to-hidden weight functions $h_{g}^i$ are supported on the $l$-th layer itself, whereas $w_{g}^i$ is supported on the $(l-1)$-th layer. 
 
 Having carried out the fusion of the first $l-1$ layers we consider the following problem to fuse the $l$-th layers:
\begin{equation}\label{eq: GWBP for rnn fusion}
		\min_{W_l, H_l, \{\Pi_l^i\}_i}  B(W_l, H_l; \{\Pi_l^i\}_i) := 
		\frac{1}{n} \sum_{i=1}^{n} \langle \mathcal{L}(W_l, W_l^i) \otimes (\Pi_{l-1}^i)^* + \alpha_H\mathcal{L}(H_l, H_l^i) \otimes \Pi_{l}^i, \Pi_{l}^i\, \rangle,
\end{equation}
where $\alpha_H$ is a hyperparameter that balances the importance of input-to-hidden weights and hidden-to-hidden weights during the fusion; we'll use $(\Pi_{l}^i)^*$ to denote an optimal $\Pi_{l}^i$. 
{We use $H_l$ and $H_l^i$ to denote the hidden-to-hidden weight function matrices of layer $N_l^{\text{new}}$ and $N_l^i$ respectively, and we let
$\mathcal{L}(H_l, H_l^i) := \big[\big(h_{jq} - h_{gs}^i\big)^2\big]_{j,g,q,s}$. $\mathcal{L}(W_l, W_l^i) $ is defined the same as in the fully connected case.} Notice that this is a GW-like barycenter problem.

We provide more detailed explanation on how to derive optimization problem \ref{eq: GWBP for rnn fusion} and adapt the GWB fusion for RNNs discussed in this section to the LSTM case in Appendix \ref{sec: GWB Fusion}.
In Section \ref{sec: experiments on GWB based fusion} we show that the models obtained when setting $\alpha_H >0$ in \eqref{eq: GWBP for rnn fusion} greatly outperform the models obtained when setting $\alpha_H=0$, justifying in this way the use of GWBs.

% (Fusion algorithm when we drop the hidden to hidden connections)

% It is possible to still carry out fusion of RNN without the need to use GWB framework simply by dropping the term in \eqref{eq: GWBP for rnn fusion} (i.e. let $\alpha_H=0$) that accounts for hidden-to-hidden connections \cite{wang2020federated}.
% By doing so, we return to the WB based fusion. The couplings $\Pi_l^i$
% can be obtained by solving minimization problem \eqref{eq: problem after plugging in cost function d_W}, where $W_l, W_l^i$ are the input-to-hidden weights of RNN at $l$-th layer. And the input-to-hidden and hidden-to-hidden weights of target model are computed the same as in GWB framework. While this procedure may seem appealing at first sight, our experiments in section \ref{sec: experiments on GWB based fusion} show that
% fused models obtained by using GWB based fusion algorithm consistently outperform the ones produced by following above procedure over different settings.

\section{Experiments}\label{section: experiment}

{\textbf{Overview:}
We present an empirical study of our proposed WB and GWB based fusion algorithms to assess its performance in comparison to other state of the art fusion methodologies and reveal new insights about the loss landscapes for different types of network architectures and datasets. We first consider the fusion of models trained on heterogeneous data distributions.
% and show that for certain combinations our fused model outperforms the base models. 
Next we present results for WB fusion of FC NNs and deep CNNs, and draw connections between workings of WB fusion and LMC of SGD solutions. 
Finally, we consider GWB fusion and present results on RNNs, LSTMs and extend the conjecture made in \cite{entezari2021role} for recurrent NNs.}

%We first present the results of model fusion for different network architectures and datasets under both WB and GWB based fusion frameworks. In particular, for WB based method, we show the results on fully connected NNs and deep convolutional models like VGG11 and $\resnet$. And for GWB based method, we present the results on RNNs and LSTMs (other relevant experiments like single shot model distillation and knowledge transfer based on our fusion method can be found in the Supplement). Secondly, we visualize the fusion results of both two frameworks on two-dimensional subspace of the loss surface of NNs. By the visualizations, we explain the mechanisms behind of fusion algorithms and make connections to the conjecture in \cite{entezari2021role}.}

{\textbf{Baselines:} 
For baselines, we consider vanilla averaging and the state-of-the-art fusion methodologies like OT fusion \cite{singh2019model} and FedMA \cite{wang2020federated}. For a fair comparison under the experimental settings of one-shot fusion we consider FedMA without the model retraining step and restrict its global model to not outgrow the base models. We refer to this as ``one-shot FedMA''. For RNNs and LSTMs, our baselines additionally include slightly modified versions of WB based fusion and OT fusion where we ignore the hidden-to-hidden connections. Other methods which require extensive training are not applicable in one-shot model aggregation settings. }
% {\red We also leave the potential applications of our fusion method in federated learning settings \cite{mcmahan2017communication} for future work.
% }}

\textbf{Base models \& General-setup:} For our experiments on FC NNs, we use $\mlp$ introduced in \cite{singh2019model}, which consists of 3 hidden layers of sizes $\{400, 200, 100\}$. Additionally, we introduce $\mlpl$ and $\mlps$ with hidden layers of size $\{800, 400, 200\}$ and $\{200, 100, 50\}$ respectively.
For deep CNNs, we use VGG11 \cite{simonyan2014very} and $\resnet$ \cite{he2016deep}. For recurrent NNs, we work with RNNs and LSTMs with one hidden layer of size $256$ and $4 \times 256$ respectively.
Hyperparameters are chosen using a validation set and final results are reported on a held out test set. More training details are provided in the Appendix. 
% {\red For all loss landscape visualizations we fuse two base models and initialize the target model using the first base model.}

{\textbf{Visualization methodology:} 
We visualize the result of fusing two pre-trained models on a two-dimensional subspace of NNs' loss landscape by using the method proposed in \cite{garipov2018loss}.
In particular, each plane is formed by all affine combinations of three weight vectors corresponding to the parameters of base model 1, base model 2 and permuted model 2 (i.e. base model 2 \textit{after} multiplying its weights by the coupling obtained by our fusion algorithm) respectively.
}

\subsection{WB fusion under heterogeneous data distributions}\label{sec: WB hetero}
\textbf{Setup:}  
We first apply WB fusion in aggregating models trained on heterogeneous data distributions which is a setting often found in federated learning where the clients have local data generated from different distributions and privacy concerns prevent data sharing among them. Here we follow the setup described in \cite{singh2019model}. To simulate heterogeneous data-split on MNIST digit classification one of the models (named A) is trained with a special skill to recognize one of the digits (eg. digit 4) that is not known to the other model, named B. Model B is trained on $90\%$ of the training data for remaining digits while model A uses the other $10\%$ data. Under this data split, we consider two settings. For the first setting, the base models are fused into a target model of the same architecture ($\mlp$). For the second setting, we consider the fusion of two small base models ($\mlps$) into a large target model ($\mlp$). This simulates the setting where clients in federated learning are constrained by memory resources to train smaller models. 
{In both cases we use model fusion to aggregate knowledge learned from the base models into a single model, a more memory-efficient strategy than its ensemble-based counterparts}. 
%When the target model is of the same architecture as the base model we randomly choose one of the base models to initialize {\red the weights of} the target model before starting the fusion. However when the target model has a different architecture the base model cannot be directly used and hence we propose an additional initialization step. In this additional step, we run one-shot model fusion on a single base model and a randomly initialized target model. The resulting target model is used as initialization for the subsequent fusion algorithm. We refer to this as distillation initialization. 
% {\red For fair comparison, we also apply the distillation initialization on OT fusion, which is not mentioned in \cite{singh2019model}. If we random initialize the target model for OT fusion (the way from the original paper), the test accuracy of the fused model obtained by OT fusion will be bad.}

\begin{figure}[!htb]
    \centering
    \begin{subfigure}{.4\textwidth}%
    \includegraphics[width=1.0\linewidth]{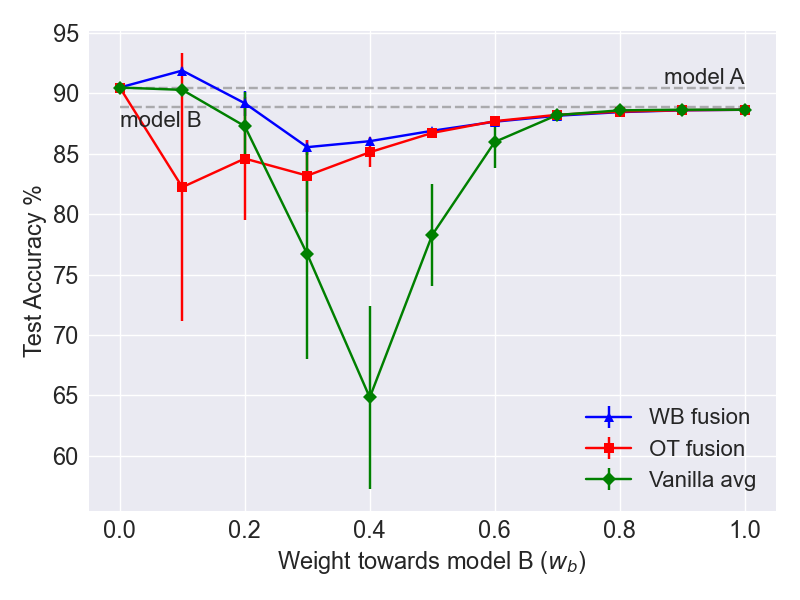}
    \end{subfigure}%
    \begin{subfigure}{.4\textwidth}
    \includegraphics[width=1.0\linewidth]{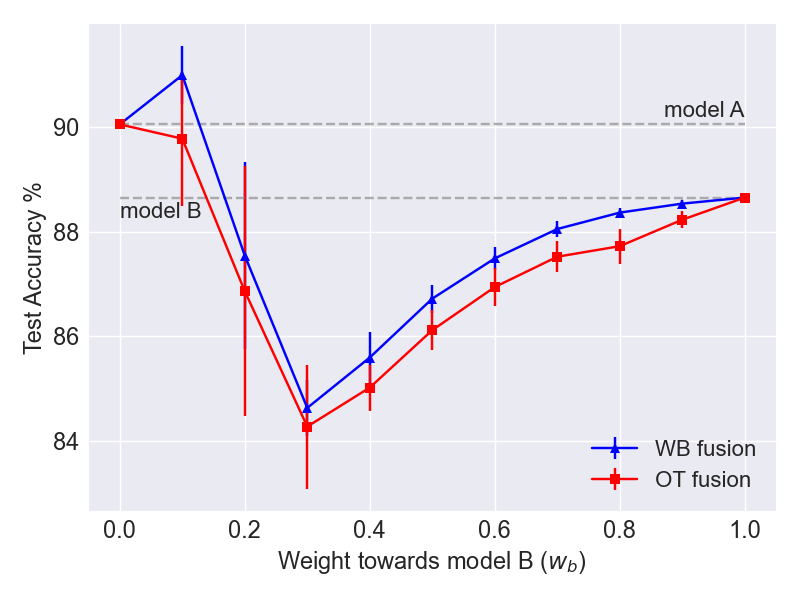}
    \end{subfigure}
\caption{\textbf{Left / Right:} Test accuracy \% for fused models when base models are trained on heterogeneous data distributions and combined with various proportions into a target model of \textbf{same / different} architecture. Some models obtained by WB fusion outperform even the base models.}
\label{fig: heterogeneous data distribution}
\end{figure}

\textbf{Quantitative results:} Figure \ref{fig: heterogeneous data distribution} shows the results of single shot fusion when different proportions of the base models are considered. We find that (a) WB fusion consistently outperforms the baselines,
%the fused models obtained from our WB algorithm consistently outperforms the baselines, 
(b) for certain combinations WB produces fused models with accuracy even better than the base model and demonstrates successful one shot knowledge aggregation. Note that for each proportion of model aggregation (x-axis),  the results are reported over multiple runs where one of the base models is randomly chosen to initialize the target model in the fusion algorithm.
% This is slightly different from \cite{singh2019model} where the target model is always initialized using model A and accounts for the differences when comparing against the plots in \cite{singh2019model}. 
We find that WB fusion is more stable against initialization as indicated by the lower variance in Figure \ref{fig: heterogeneous data distribution}. For fusion into different architectures vanilla averaging is not applicable, and we do not include ``one-shot FedMA'' for comparison here since it is not clear how to assign different proportions to base models in FedMA, or to specify a target architecture different from the base models.

\subsection{WB fusion under homogeneous data distributions and connections to LMC}\label{sec: WB homo}

{\textbf{Setup:} 
In this section we perform WB fusion for various models and architectures, and provide loss landscape visualizations which reveal workings of the fusion algorithm and shed light on linear model connectivity (LMC) of SGD solutions after applying appropriate permutations. We first consider fusion of FC NNs on the MNIST dataset \cite{deng2012mnist} and train $\mlp$ following \cite{singh2019model}. For this we consider two different settings. In the first setting, the target model has the same architecture as the base models. For the second one, we fuse the base models into a larger model $\mlpl$. As noted before, the latter scenario is relevant for federated learning, given the limitations of memory and computational resources on edge devices. Next, we consider fusion of deep CNNs like VGG11, $\resnet$ trained on CIFAR10 dataset \cite{krizhevsky2009learning}. For all these cases, we fuse 2 trained models initialized differently. For the skip-connection and fusion into different architectures, FedMA is not directly applicable and hence not considered for comparisons.}

{\textbf{Quantitative results:} Table \ref{table: WB fusion comparison} contains the results of fusion for FC NNs and deep CNNs. We find that (a) WB fusion produces models at par or outperforms other fusion methods for all considered model types and datasets, (b) for fusion into different architectures and ResNets, we find that WB fusion is more effective and robust.}
% {\red We attribute these to the fact that, in comparison to OT fusion, our algorithm iterates until convergence (see more details in the Appendix). }}
%{\red contains the results of single shot model fusion for fully connected NNs and deep CNNs. 
%(a) We find that WB fusion outperforms 
%other fusion types irrespective of model types and datasets. (b) When considering the hard settings like fusing into different architectures or the base models being deep neural networks, the WB fusion is more robust to the initialization of target model. This is because, compared to OT fusion, our fusion algorithm iterates until convergence, which eliminates the randomness from the initialization.}

\begin{table}[!htb]
	\caption{Performance comparison (Test accuracy $\pm$ standard deviation \%) of different fusion algorithms under various network architectures and datasets. ``BASE'' means initializing target model with one of the base models. For each case, the target model obtained by WB fusion gets the highest test accuracy and smallest standard deviation.}
	\label{table: WB fusion comparison}
	\centering
	\renewcommand\arraystretch{1.5}
	\resizebox{0.7\textwidth}{!}{
		\begin{sc}
			\begin{tabular}{@{}cccccc@{}}
				%\toprule
				& \multicolumn{2}{c}{MNIST} & \phantom{ab}& \multicolumn{2}{c}{CIFAR10}  \\ 
				\cmidrule{2-3} \cmidrule{5-6} 
				&  $\mlp$/base & $\mlpl$ & & VGG11/base & $\resnet$/base  \\ 
				\midrule
				Base Model Avg & $98.31 \pm 0.02$ & - & & $90.14 \pm 0.19$ & $91.56 \pm 0.34$ \\
				\midrule[0.05pt]
				Vanilla Avg & $86.50 \pm 4.60$ & - & & $30.82 \pm 4.49$ & $20.56 \pm 3.90$ \\
				One-shot FedMA & $97.89 \pm 0.10$ & - & & $\bm{85.42 \pm 1.01}$ & - \\
				OT & $97.84 \pm 0.12$ &  $91.53 \pm 2.64$& & $85.39 \pm 0.93$ & $71.37 \pm 6.53$ \\
				WB & $97.92 \pm 0.12$&  $\bm{94.93 \pm 1.18}$ & & $85.39 \pm 0.93$ &  $\bm{73.75 \pm 4.39}$ \\
				\bottomrule
			\end{tabular}
		\end{sc}
	}
\end{table}

\begin{figure}[!htb]
    \centering
    \begin{subfigure}{.33\textwidth}%
    \includegraphics[width=1.0\linewidth]{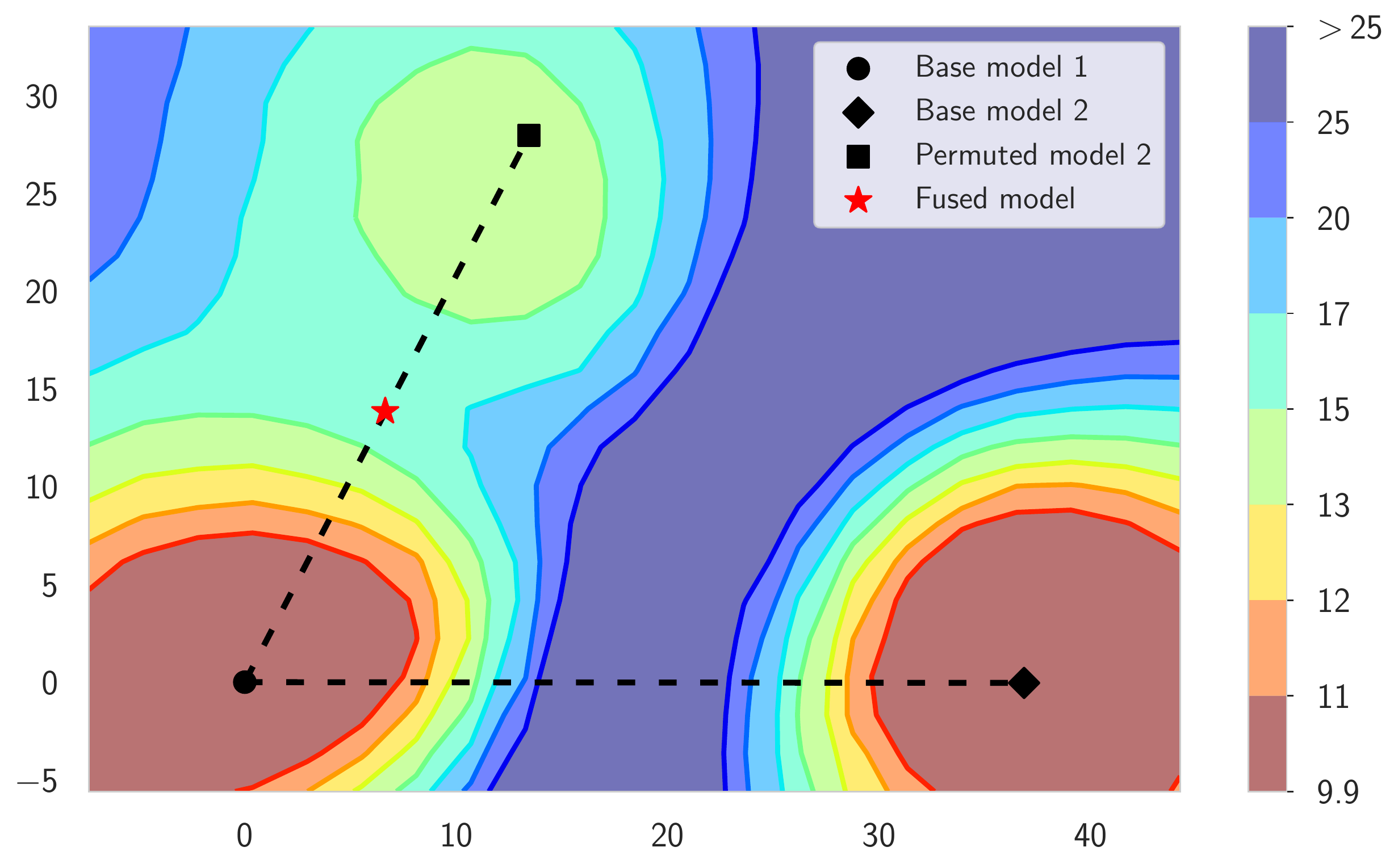}
    \end{subfigure}%
    \begin{subfigure}{.33\textwidth}%
    \includegraphics[width=1.0\linewidth]{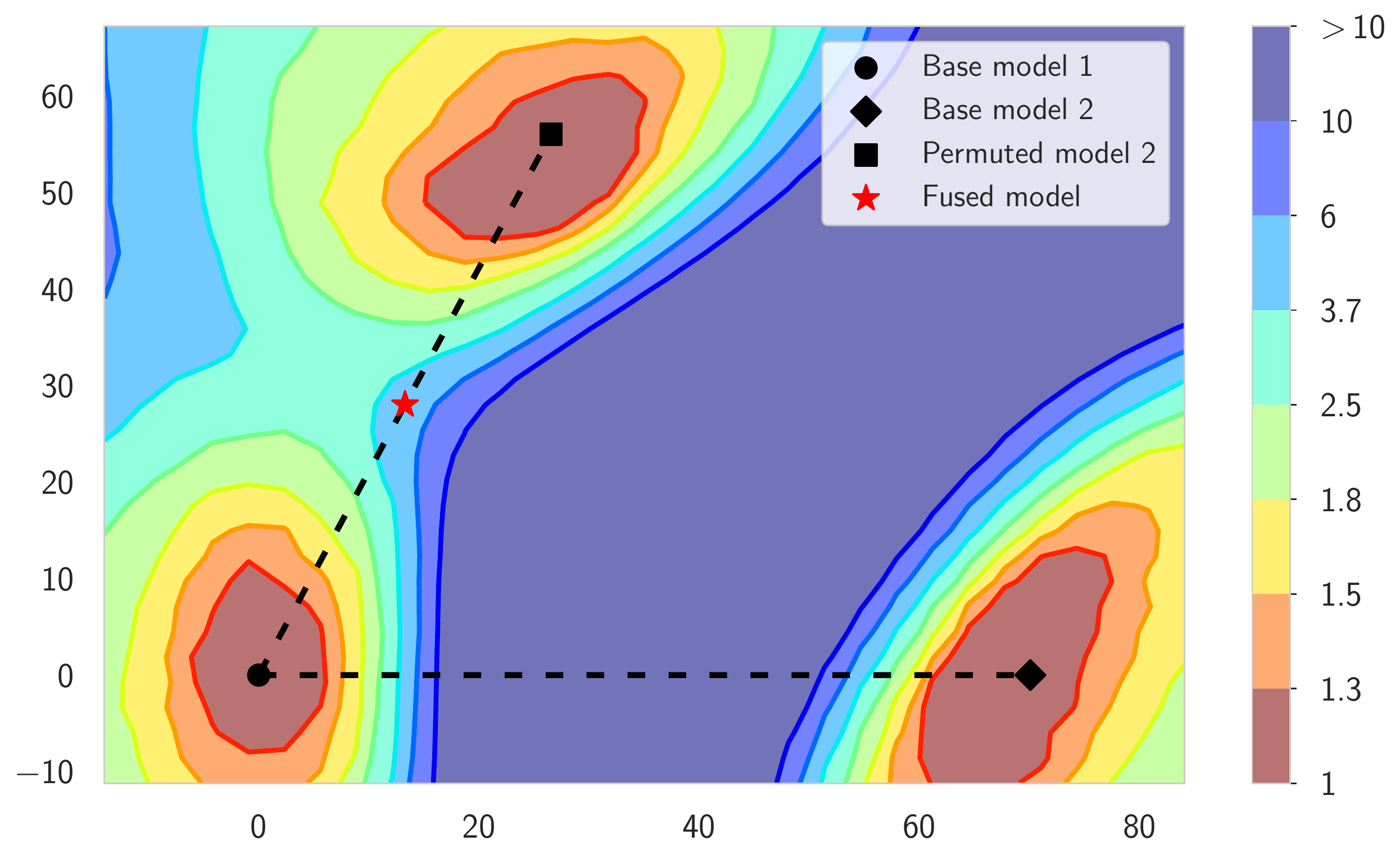}
    \end{subfigure}%
    \begin{subfigure}{.33\textwidth}%
    \includegraphics[width=1.0\linewidth]{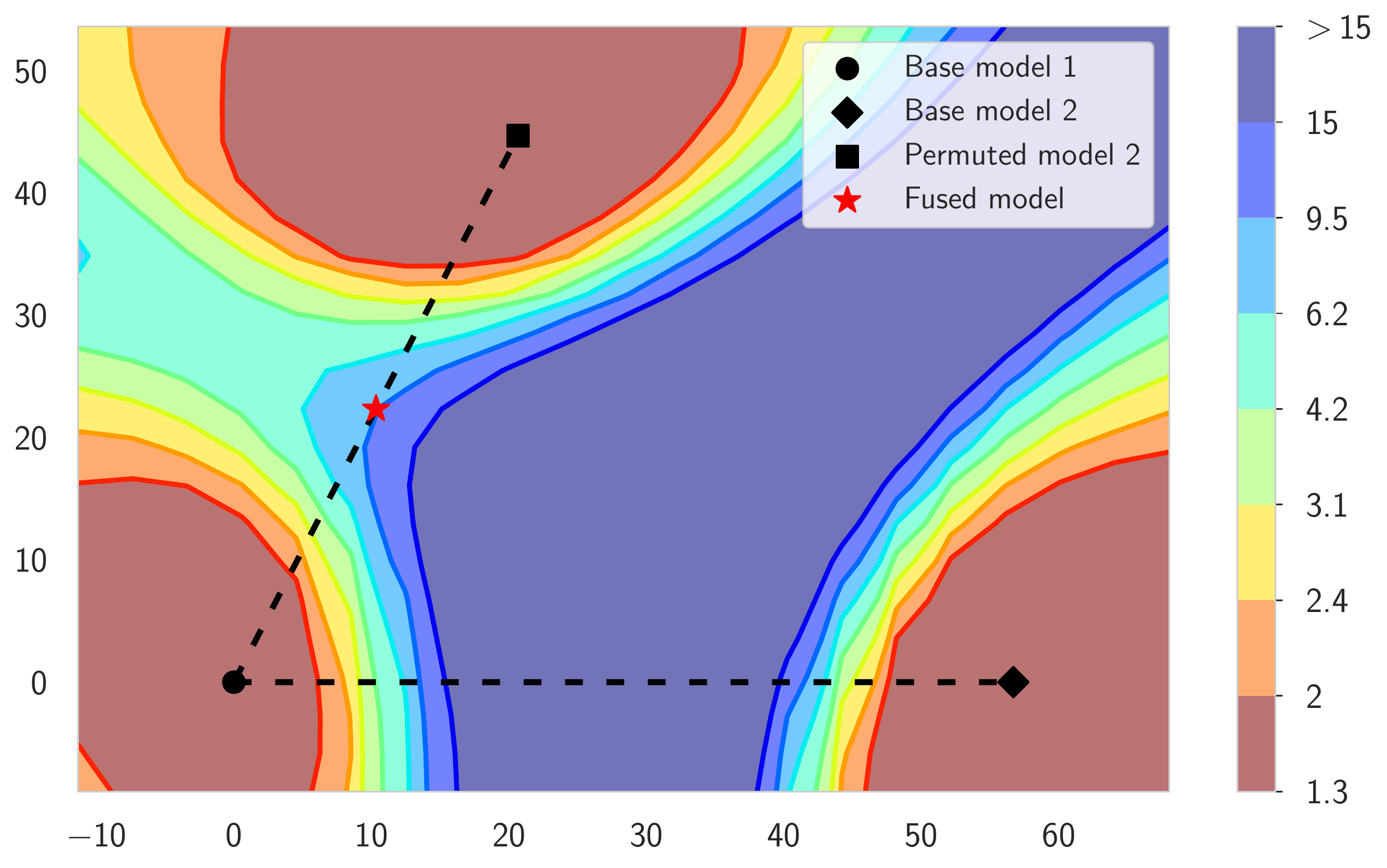}
    \end{subfigure}
\caption{Visualizations of the fusion results on the test error surface, which is a function of network weights in a two-dimensional subspace, for different models and datasets. \textbf{Left:} Fusion of two VGG11 models trained on CIFAR10 dataset using WB framework. \textbf{Middle:} Fusion of two LSTM models trained on MNIST dataset . \textbf{Right:} Fusion of two LSTM models trained on DBpedia dataset. We can observe that in all these cases the basins of permuted model $2$ (obtained by multiplying the weights of base model $2$ by the found coupling) and base model 1
lie close to each other and are separated by a low energy barrier.}
% \textit{after} multiplying the couplings found by our fusion algorithm, base model 2 will end up lying in the basin that close to one model 1 lands in.}
\label{fig: visualization of fusion results for different types of models and datasets}
\end{figure}

\textbf{Visualizations:} 
Figure \ref{fig: visualization of FCNN training on MNIST and illustration of FCNN} (left) contains the visualization of fusing two $\mlp$ trained on MNIST dataset under WB framework and Figure \ref{fig: visualization of fusion results for different types of models and datasets} (left) contains the fusion result of WB fusion of two VGG11 models trained on CIFAR10. We find that (a) the couplings obtained in WB fusion (refer to equation \ref{eq: problem after plugging in cost function d_W}) between the layers of target model and base models are sparse, i.e. they are almost permutations; (b) the basins of the permuted model 2 (obtained by multiplying the weights of base model 2 by the found couplings) and base model 1 lie close to each other and are separated by a low energy barrier. These visualizations thus provide new empirical evidence in support of the conjecture made in \cite{entezari2021role}. They also shed light on the workings on WB fusion algorithm. In particular, equation \ref{eq: update formula for W_l} can be interpreted as coordinate-wise averaging of the permuted models. Since permuted models land in basins that are separated by low energy barriers, their linear interpolation gives a good fused model.

\subsection{GWB fusion for recurrent neural networks}\label{sec: experiments on GWB based fusion}
{\textbf{Setup:}
In this section, we consider the fusion of NNs like RNNs and LSTMs on sequence based tasks. 
We use 4 different datasets for this setting: 
i) MNIST \cite{deng2012mnist}: Images of $28\times 28$ dimensions are interpreted as $28$ length sequences of vectors $\in \mathbb{R}^{28}$; 
ii) SST-2 \cite{socher2013recursive}: Binary classification task of predicting positive and negative phrases; 
iii) AGNEWS \cite{zhang2015character}: Corpus of news articles from 4 classes;
and iv) DBpedia \cite{zhang2015character}: Ontology classification dataset containing 14 non-overlapping classes. 
% The base models are RNNs and LSTMs with one hidden layer and 
% hidden dimensions of size $256$ and $4\times 256$ respectively. 
For the NLP tasks, we use pre-trained GloVe embeddings \cite{pennington2014glove} of dimensions $100$ and $50$
for RNNs and LSTMs respectively. 
The embedding layer is not updated during the model training.
We set the target model to have the same architecture as the base models.
}
%{\red \textbf{Setup:} In this section, we show the importance of utilizing the hidden-to-hidden weights of RNN during the fusion by comparing the performance of GWB based fusion algorithm with other fusion types. We consider $4$ different tasks for the experiments on RNN and LSTM models. Specifically, the $4$ datasets are: i) MNIST dataset; ii) AGNEWS dataset ..; iii) DBpedia dataset ..; iv) SST dataset .. . In general, the base models are RNNs and LSTMs with one hidden layer. The hidden dimensions of RNN and LSTM models are $256$ and $4 \times 256$ respectively. For the image task (MNIST dataset), the input dimension and the time step of RNN and LSTM are set to be $28$, which fits the size of the digital image in MNIST; for the NLP tasks (AGNEWS, DBpedia, SST datasets), we consider RNN and LSTM models consisting of a pre-trained embedding layer (embedding vary for different datasets, see the detailed settings of both image and NLP tasks in the Supplement). We set target model to have the same architecture as the base models. And when conducting fusion on NLP datasets, we assume that the target model shares the same pre-trained embedding with base models.}

{\textbf{Quantitative results:} Table \ref{table: GWB fusion comparison} contains the result of fusion for various datasets and model architectures. We find that 
(a) our GWB framework outperforms other fusion algorithms for each combination of model type and dataset, 
which highlights the importance of using hidden-to-hidden connections for the fusion of recurrent NNs; 
(b) the accuracy gains for GWB over WB is different for different tasks, 
which indicates that relative importance of hidden-to-hidden connections is task dependent; 
(c) the accuracy of fused model is higher for LSTMs in comparison to RNNs,
which we attribute to the fact that LSTMs have four hidden states 
and thus four input-to-hidden and hidden-to-hidden weight matrices. 
More information for each hidden node allows the algorithm to uncover better couplings.
Our results in (a) and (b) show the usefulness of hyperparameter $\alpha_H$ (set between $[1,20]$) from \eqref{eq: GWBP for rnn fusion}
in balancing the relative importance of hidden-to-hidden weights. 
% In our experiments,
% we find that setting $\alpha_H \in [1, 20]$ 
% results in better fused model for GWB. 
}

\begin{table}[htb]
	\caption{Performance comparison (Test accuracy $\pm$ standard deviation \%) of different fusion algorithms under various network architectures and datasets. For each case, target model obtained by GWB fusion reaches the highest test accuracy and small standard deviation.}
	\label{table: GWB fusion comparison}
	\centering
	\renewcommand\arraystretch{1.5}
	%	\setlength{\abovecaptionskip}{0pt}%    
	%	\setlength{\belowcaptionskip}{10pt}%
	%\resizebox{0.5\textwidth}{0.09\textwidth}{
	\resizebox{1.0\textwidth}{!}{
		\begin{sc}
			\begin{tabular}{@{}cccccccccccc@{}}
				%\toprule
				& \multicolumn{2}{c}{MNIST} & \phantom{ab}& 
				\multicolumn{2}{c}{AGNEWS} &
				\phantom{ab}&
				\multicolumn{2}{c}{DBpedia} &
				\phantom{ab}&
				\multicolumn{2}{c}{SST-2}\\ 
				\cmidrule{2-3} \cmidrule{5-6} \cmidrule{8-9} \cmidrule{11-12}
				&  RNN & LSTM&  & RNN & LSTM & & RNN & LSTM & & RNN & LSTM \\ 
				\midrule
				Base Model Avg & $96.68 \pm 0.29$ & $98.99 \pm 0.09$ &  & $88.68 \pm 0.12$& $92.38 \pm 0.17$& & $97.12 \pm 0.21$ & $98.62 \pm 0.11$ & & $87.32 \pm 1.03$& $90.31 \pm 0.27$\\
				\midrule[0.05pt]
				Vanilla Avg & $28.54 \pm 10.70$ & $31.92 \pm 4.86$ &  & $40.77 \pm 4.94$& $74.01 \pm 3.89$& & $30.95 \pm 4.53$& $50.93 \pm 2.17$ & & $73.91 \pm 2.73$& $74.25 \pm 1.92$\\
				OT & $36.78 \pm 14.13$ & $68.33 \pm 7.07$ &  & $53.05 \pm 4.30$& $86.19 \pm 2.14$& & $37.91 \pm 4.86$& $77.95 \pm 3.20$ & & $78.92 \pm 2.97$& $82.13 \pm 0.60$\\
				One-shot FedMA & $34.16 \pm 7.26$ & $66.98 \pm 5.17$ & & 
				$55.78 \pm 3.64$ & $86.30 \pm 2.40$ & &
				$42.16 \pm 6.24$ & $81.81 \pm 3.29$ & &
				$79.17 \pm 2.27$ & $82.53 \pm 1.01$ \\
				WB  & $29.41 \pm 7.05$ & $67.66 \pm 6.27$ &  & $55.63 \pm 4.18$ &  $86.25 \pm 2.37$ &  & $42.52 \pm 6.26$& $82.57 \pm 3.55$ & & $79.57 \pm 2.36$ & $82.87 \pm 1.09$\\
				GWB & $\bm{81.39 \pm 2.97}$ & $\bm{93.27 \pm 1.86}$ & & $\bm{61.01 \pm 3.87}$ & $\bm{87.96 \pm 0.91}$ & & $\bm{55.15 \pm 5.97}$& $\bm{87.50 \pm 2.89}$ & & $\bm{82.60 \pm 1.05}$ & $\bm{84.04 \pm 0.77}$\\
				\bottomrule
			\end{tabular}
		\end{sc}
	}
\end{table}

\textbf{Visualizations:} Figure \ref{fig: visualization of fusion results for different types of models and datasets} (middle, right) 
contains visualization of fusing LSTM models under the GWB framework. 
{As noted for the FC NNs and deep CNNs visualizations, we find that (a) the couplings found by GWB fusion algorithm 
are sparse, and (b) these couplings map different local minima into neighboring basins 
that are separated by low energy barriers. }
This empirical evidence suggests that the original conjecture in \cite{entezari2021role}
can be extended to richer network architectures and tasks (RNNs and LSTMs on NLP datasets).

\section{{Conclusion}}

{In this paper we have proposed neural network fusion algorithms that are based on the concept of Wasserstein/Gromov-Wasserstein barycenter. Our fusion algorithms allow us to aggregate models within a variety of NN architectures, including RNN and LSTM. Through extensive experimentation we: 1) illustrated the strengths of our algorithms 2) provided new empirical evidence backing recent conjectures about the linear mode connectivity of different neural networks with architectures such as RNN or LSTM and for different imaging and NLP datasets.} 

\textbf{Limitations and future work:} NNs with ReLU activation are also scale-invariant across the layers which is currently not handled in our cost functions. 
Although the empirical evidence in \cite{du2018algorithmic,entezari2021role} suggests that the models trained on same datasets using SGD converges to solutions with more balanced weights, it might be the case that for certain heterogeneous settings the weights across models become less balanced. For future work we would like to explore fusion using scale-invariant cost functions and apply WB/GWB fusion algorithms to federated learning.

% Although our WB and GWB frameworks consistently outperform other fusion algorithms, the test accuracy of fused model is relatively low in comparison to the base model for cases like $\resnet$ trained on CIFAR10 dataset or RNN trained on AGNEWS and DBpedia datasets. This might be attributed to the fact that we only conduct single shot model fusion without using any data. This motivates us to explore the use of our fusion algorithms within federated learning, where aggregation occurs while training. We leave this for future work. 

% \begin{remark}\label{remark: difference between WB and OT}
% % The WB barycenter problems underlying our algorithm provide a principled way to interpret the fusion problem. 
% The update proposed in Step 1 and 2 is closely related to the OT fusion algorithm from \cite{singh2019model}, yet with some differences highlighted in the Supplement. At a higher level, by explicitly writing the fusion problem as a minimization problem (a barycenter problem), as we do here, it becomes clear that one should iterate Steps 1 and 2 until convergence, rather than stopping after one iteration as in OT fusion. In our numerical experiments we illustrate the significance of iterating multiple times, particularly when the weights of the target model are initialized randomly.
% \end{remark}

\bibliography{ms}
\bibliographystyle{iclr2023_conference}

\appendix

\section{Related works}
\subsection{Other model fusion algorithms}\label{sec:Otherfusion}
Different from the post-processing strategies proposed in \cite{singh2019model,yurochkin2019bayesian,wang2020federated,liu2022deep}, which propose aligning different models \textit{after} training, works like \cite{yu2021fed2,li2022federated} solve the alignment problem \textit{during} the training process of local models. In particular, the work \cite{yu2021fed2} proposes to align features during training by separating features into different groups; \cite{li2022federated}, on the other hand, proposes to break permutation invariance by adding position encodings during training. With any of these approaches coordinate-wise (vanilla) averaging becomes sensible and local models can get aligned during local updates using direct averaging. The work \cite{claici2020model} considers a general model fusion problem (e.g. topic models, not restricted to NN) and takes a Bayesian approach. In their framework, the parameters specifying the target global model are obtained by minimizing the sum of KL distances between ``posterior distributions'' of global and local models. \cite{chen2020fedbe} also takes a Bayesian perspective: local models are used to estimate an ensemble of Gaussian or Dirichlet distributions and the output ensemble is used to generate the global model using knowledge distillation. In \cite{nguyen2021model}, the problem of fusing NNs with different number of layers is considered. Their idea is to first apply dynamic programming to find one-to-one cross-layer alignments. Based on these alignments, the number of layers of local NNs gets balanced, and they can then reduce their fusion problem to one where existing layer-wise fusion methods can be used.

\subsection{ Computation of Wasserstein barycenters}\label{sec: computational OT}

	Solving Wasserstein barycenter problems (WBP)
	has become computationally more feasible because of recent substantial advancements in OT algorithms.
	This development started with the work \cite{cuturi2013sinkhorn} where they proposed to add an entropic regularization term in the transport problem and use the Sinkhorn algorithm \cite{sinkhorn1967diagonal,knopp1968note} to solve it. \cite{cuturi2014fast} and \cite{benamou2015iterative} extended the entropic regularization idea to the barycenter problem and proposed to solve it through sub-gradient descent and Bregman projection iterations respectively. In \cite{claici2018stochastic}, the authors proposed to alternatively update the measure support and the weights of the barycenter by using stochastic algorithms. The work \cite{peyre2016gromov} proposed to use projected gradient descent to solve the minimization problem corresponding to the entropic Gromov-Wasserstein (GW) discrepancy, which is shown to be equivalent to solve a entropy-regularized OT problem at each iteration. \cite{xu2019scalable,xu2019gromov} replaced the entropy regularizer with KL-divergence in the GW discrepancy and similarly solve it with proximal point methods. Both of the above methods are applied to solve GWBP. Thanks to the development of these computational techniques, Wasserstein barycenters have been applied to various machine learning problems such as clustering \cite{ho2017multilevel},
	Bayesian inference \cite{srivastava2015wasp,srivastava2018scalable}, domain adaptation \cite{redko2019optimal}, ensemble learning \cite{dognin2019wasserstein}, 
	among others. Here we apply some of these computational tools for NN fusion.

\section{Computational optimal transport}
In this section, we review the computational optimal transport methods used in our paper.

\subsection{Entropic Optimal Transport}
Let $p \in \Sigma_{N_1}$ and $q \in \Sigma_{N_2}$ be two histograms and let $C \in \mathbb{R}_+^{N_1 \times N_2}$ be a cost matrix, where $C_{ij}$ represents the transportation cost between positions indexed by $i$ and $j$. Define the solution of entropically-regularized optimal transport between $p$ and $q$ as
\begin{equation}
	\mathcal{T}(C, p, q) := \argmin_{\Pi \in \Gamma(p, q)} \, \langle C, \Pi\rangle - \varepsilon H(\Pi),
\end{equation}
where $H(\Pi) := -\sum_{i, j=1}^{N} \Pi_{ij}(\log \Pi_{ij} -1)$ is the entropy of $\Pi$. It can be shown that the solution to this problem reads $\mathcal{T}(C, p, q) = \text{diag}(a)K\text{diag}(b)$, where $K := e^{-\frac{c}{\varepsilon}} \in \mathbb{R}_+^{N_1 \times N_2}$ is the so-called Gibbs kernel associated to $c$, and $(a, b) \in \mathbb{R}_+^{N_1} \times \mathbb{R}_+^{N_2}$ can be computed using Sinkhorn iterations \cite{cuturi2013sinkhorn}
\begin{equation}\label{eq: sinkhorn iterations}
	a \leftarrow \frac{p}{Kb} \quad \text{and} \quad b \leftarrow \frac{q}{K^{\top}a},
\end{equation}
where here $\frac{\cdot}{\cdot}$ denotes elementwise division.

\subsection{Gromov-Wasserstein distance and Gromov-Wasserstein barycenters}
\label{sec:GWBar}

The Gromov-Wasserstein problem  \cite{memoli2011gromov,peyre2016gromov} is a variant of the OT problem introduced with the purpose of comparing different spaces when each of them is endowed with a base probability distribution and a notion of similarity (or dissimilarity) between pairs of its elements.  
% Optimal transport needs a ground cost $c$ to compare the probability measures $\mu$ and $\nu$ and thus cannot be used if the supports of those measures are not defined on the same underlying space. 
For two matrices $C \in \mathbb{R}^{n_1 \times n_1}$ and $\bar{C} \in \mathbb{R}^{n_2 \times n_2}$ representing the similarity between pairs of points in the support of $\mu = \sum_{i=1}^{n_1}a_i \delta_{x_i}$ and $\nu = \sum_{j=1}^{n_2} b_j \delta_{y_j}$ respectively, we define the Gromov-Wasserstein distance between the two measured similarity matrices $(C, a) \in \mathbb{R}^{n_1 \times n_1} \times \Sigma_{n_1}$ and $(\bar{C}, b) \in \mathbb{R}^{n_2 \times n_2} \times \Sigma_{n_2}$ as follows:
\begin{equation}\label{eq: GW-distance}
	GW((C, a), (\bar{C}, b)) := \min_{\Pi \in \Gamma(a, b)} \langle \mathcal{L}(C, \bar{C}) \otimes \Pi, \Pi\rangle,
\end{equation} 
where $\mathcal{L}(C, \bar{C}) := \big[L(C_{ik}, \bar{C}_{jl})\big]_{i,j,k,l}$ and $L$ is some loss function to account for the misfit between the similarity matrices. In direct analogy with the barycenter problem in optimal transport, we define the Gromov-Wasserstein barycenter problem (GWBP) for the measured similarity matrices $\{(C_i, b_i)\}_{i=1}^{n}$ using a Fr\'{e}chet mean formulation:
\begin{equation}
	\min_{C \in \mathbb{R}^{m \times m},\, \{\Pi^i\}_i} \,\,\frac{1}{n} \sum_{i=1}^{n} GW((C, a), (C_i, b_i)) = 
     \frac{1}{n} \sum_{i=1}^{n} \langle \mathcal{L}(C, C_i) \otimes \Pi^i, \Pi^i\rangle,
\end{equation}
where we assume $a \in \Sigma_m$ to be known. A minimizer of this problem is called Gromov-Wasserstein barycenter (GWB) of  the measured similarity matrices $\{(C_i, b_i)\}_{i=1}^{n}$ (for fixed $a$). We used the concept of GWB to define fusion algorithms for RNN and LSTM—see Section \ref{sec:GWFusion}.

\subsection{Entropic Gromov-Wasserstein distance}
In order to solve the minimization problem (\ref{eq: GW-distance}), consider the following entropic approximation of the initial GW formulation:
\begin{equation}
	GW_{\varepsilon}((C, p), (\bar{C}, q)) := \min_{\Pi \in \Gamma(p, q)} \mathcal{E}_{C, \bar{C}}(\Pi) - \varepsilon H(\Pi).
\end{equation}
In \cite{peyre2016gromov}, the authors propose to use projected gradient descent to solve this non-convex optimization problem, where both the gradient step and the projection are computed according to the Kullback-Leibler (KL) divergence. Iterations of the corresponding algorithm are given by
\begin{equation}\label{eq: iter of solve GW (Proj)}
T \leftarrow \text{Proj}_{\Gamma(p,q)}^{\text{KL}} \big(T \odot e^{-\tau (\nabla \mathcal{E}_{C, \bar{C}}(\Pi) - \varepsilon \nabla H(\Pi))} \big),
\end{equation}
where $\tau >0$ is a small enough step size, and the KL projector of any matrix $K$ is defined as
\begin{equation}
	\text{Proj}_{\Gamma(p,q)}^{\text{KL}}  := \argmin_{\Pi' \in \Gamma(p, q)} KL(\Pi' | K).
\end{equation}
\begin{prop}[Proposition 2 in \cite{peyre2016gromov}]
	In the special case $\tau = \frac{1}{\varepsilon}$, iteration (\ref{eq: iter of solve GW (Proj)}) reads
	\begin{equation}\label{eq: iter of solve GW (EOT)}
		\Pi \leftarrow \mathcal{T}(\mathcal{L}(C, \bar{C}) \otimes \Pi, p, q).
	\end{equation}  
\end{prop}
% {\red \begin{remark}
% Thinking of iteration step (\ref{eq: iter of solve GW (EOT)}) from the perspective of optimal transport, the idea of above algorithm is that, at each iteration, fix the current coupling $\Pi$, which relaxes the quadratic problem (GW problem) (\ref{eq: GW-distance}) into a linear problem (OT problem), and then solve it using standard Sinkhorn algorithm. Our idea of fixing the support $\gamma_{l}$ when defining the cost $c_{l}(z_{l,j}, v_{l, h}^i)$ is similar to the motivation of the relaxation above, since given $\gamma_{l}$ fixed, the coupling $\Pi_{l}^i$ between $\gamma_{l}$ and $\gamma_{l}^i$ would be fixed , and at this stage $d_{l, H}^2(\cdot, \cdot)$ (defined based on $\Pi_{l}^i$) would be well-defined.
% \end{remark}}

\section{Details on WB fusion using $TL^p$ formalism}\label{sec: detailed WB fusion}

\subsection{$TL^p$ space}
\label{sec:TLpsub}
We first review concepts from $TL^p$ spaces which we use to motivate our interpretations of nodes in NNs.
Let $\mu \in \mathcal{M}_1^+(\mathcal{X})$ be a probability measure and let 
$L^p(\mu; \mathbb{R}^r) := \big\{f: \mathcal{X} \rightarrow \mathbb{R}^r \, |\, \int_{\mathcal{X}} \|f\|_p^p d\mu(x) < \infty\big\}$. \nc We define the $TL^p$ space  associated to $\mathcal{X}$ (see \cite{GarciaTrillos2015,thorpe2017transportation} and references within) as
\begin{equation}
	TL^p(\mathcal{X}) := \big\{(\mu, f)\, |\, \mu \in \mathcal{M}_1^+(\mathcal{X}), f \in L^p(\mu; \mathbb{R}^r)\big\}.
\end{equation}
{The $TL^p$ distance for pairs $(\mu_1, f_1), (\mu_2, f_2) \in TL^p(\mathcal{X})$ is defined by
\begin{equation}
d^p_{TL^p}\big((\mu_1, f_1), (\mu_2, f_2)\big) := \min_{\pi \in \Gamma(\mu_1, \mu_2)}  \int_{\mathcal{X} \times \mathcal{X}} c(x_1, x_2; f_1, f_2) d\pi,
\end{equation}
where the cost function is chosen to be $c(x_1, x_2; f_1, f_2) := |x_1 - x_2|^p_p + |f_1(x_1) - f_2(x_2)|^p_p$. }

In this paper we will mainly consider the case $p=2$. The idea behind 
the above construction of $TL^p$ space
is to provide a common space where it is possible to compare functions that have different supports. In other words, suppose that $f_1: \mathcal{X}_1 \rightarrow \mathbb{R} $ and $f_2: \mathcal{X}_2 \rightarrow \mathbb{R} $ are two functions where $\mathcal{X}_1, \mathcal{X}_2$ are subsets of $\mathcal{X}$. While a direct comparison between $f_1$ and $ f_2$ would be possible if $\mathcal{X}_1$ and $\mathcal{X}_2$ were the same, it is less clear how to compare them when their domains are different. In the $TL^p$ interpretation 
we think of $f_1$ as a pair $(\mu_1, f_1)$ where $\mu_1$ is a probability measure supported on $\mathcal{X}_1$ and $f_2$ is interpreted similarly. Then, after finding a suitable coupling between the measures $\mu_1$ and $\mu_2$, one can couple the functions $f_1$ and $f_2$ and use a direct $L^2$-comparison. The notation $TL^p$ suggests the use of an $L^p$ comparison after solving an optimal transport problem.
This idea has been used for the task of domain adaptation \cite{OTDomainAdaptation}.

\begin{figure}[htb]
	\centering
	\includegraphics[width=0.5\textwidth]{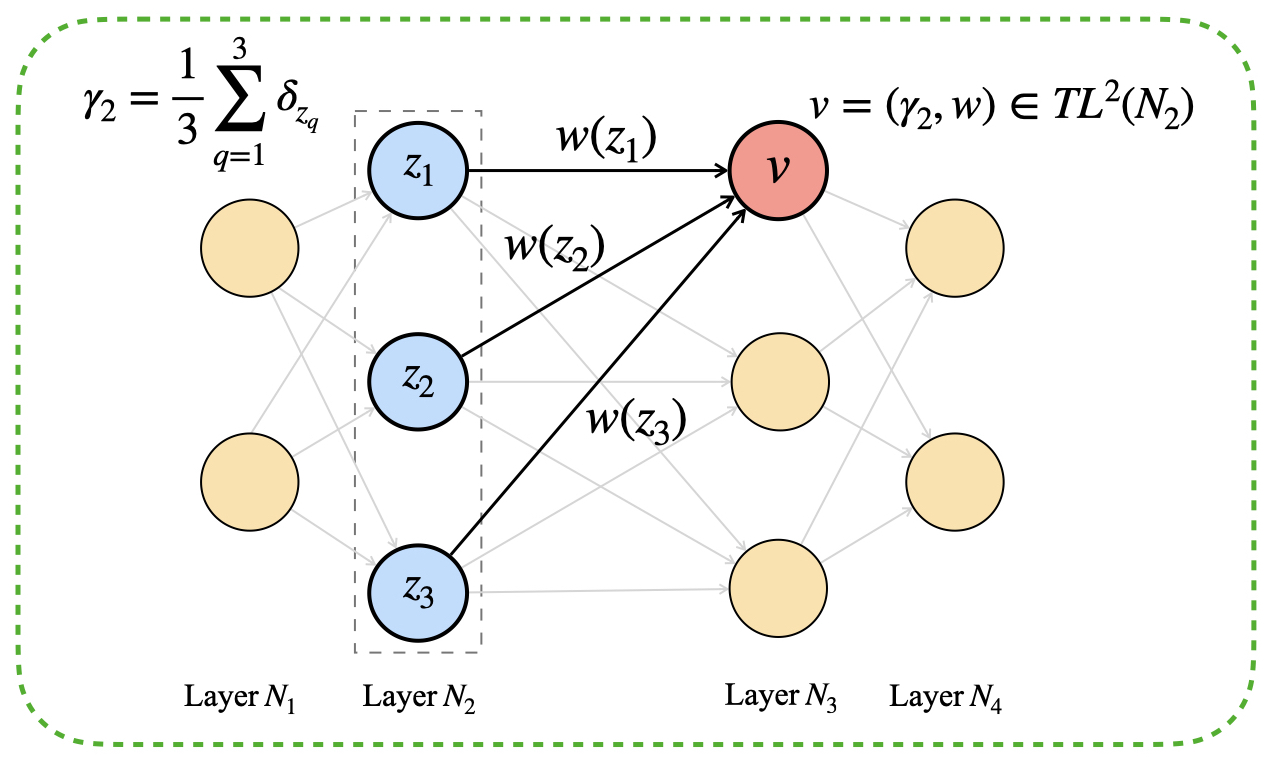}
	\caption{Following our definitions, node $v = (\gamma_2, w) \in TL^2(N_2)$, where $\gamma_2$ is a probability measure on layer $N_2$ and $w: N_2 \rightarrow \mathbb{R}$ is the weight function corresponding to node $v$. For example, $w(z_2)$ is the true parameter (weight) between nodes $v$ and $z_2$.}
	\label{fig: FC nn example with TLp}
\end{figure}

\subsection{Nested definition of fully connected NNs using $TL^p$ formalism}
{ 
We now provide detailed explanations for our interpretations of a node in a neural network based on $TL^p$ formalism. 
Intuitively, we need to construct a common space that allows us to talk about the ``distance'' between neurons in different neural networks (at the same layer). 
Therefore, inspired by the idea of $TL^p$ space, we interpret node $v$ in the $l$-th layer $N_l$ (for $l>1$) as an element in $TL^2(N_{l-1})$, that is, as a function on the domain $N_{l-1}$ (previous layer) coupled with a probability measure. 
In particular, the node $v$ is interpreted as $v := (\gamma_{l-1}, w) $, 
where $\gamma_{l-1}$ is a measure on the previous layer $N_{l-1}$ 
and $w$ represents the collection of weights between the node $v$ and the nodes in previous layer $N_{l-1}$, which can be interpreted as a function $ w : N_{l-1} \rightarrow \R$. 
An illustration of the above interpretation for $v$ is shown in Figure \ref{fig: FC nn example with TLp}. 

Next, to construct a common space using the above interpretations, we let $\mathcal{N}_l := TL^2(\mathcal{N}_{l-1})$ to be the collection of all the neurons on the $l$-th layer for $l \geq 2$ (over different neural networks). 
A simple inductive argument shows that $N_l^i \subseteq \mathcal{N}_l$ for all $i$ and all $l \geq 2$.  We can now embed the nodes on the $l$-th layers of different input models into a single common space $\mathcal{N}_l$, and in turn we can define a notion of ``distance'' to compare neurons in different neural networks (see subsection \ref{sec: cost function}).
For the first layer, we simply let $\mathcal{N}_1 := N_1^1$, since we assume all the base models have the same input layer.}

\subsection{WB fusion algorithm}\label{sec: WB fusion alg}
The WB fusion algorithm is summarized in Algorithm \ref{alg: model fusion}.

\begin{algorithm}[!ht]
	\caption{WB Fusion}
	\label{alg: model fusion}
	\begin{algorithmic}
		\REQUIRE 
		Neural networks $N^1, \cdots, N^n$ ($m$ layers);\\
		Number of nodes $k_l$ for layer $N_{l}^{\text{new}}$,  for $l =2 , \cdots, m$;\\
		Initialized weight functions $W_2, \cdots, W_{m}$;\\
		Set $(\Pi_1^i)^*$ to be the identity matrix in $\R^{k_1}$ for all $i=1, \dots, n$. Set also $\gamma_1= \gamma_1^1$;
		\FOR{$l = 2, \cdots, m$}
		%		\STATE Choose $\mu_l = \gamma_{l-1}$ and define $d_W^2$ based on couplings $(\Pi_{l-1}^i)^*$'s;\\
		\REPEAT
		\STATE \textbf{Step 1:} for Wasserstein barycenter problem (\ref{eq: problem after plugging in cost function d_W}), fix $W_l$ and obtain couplings $\Pi_l^i$ by solving $n$ independent OT problems;\\
		\STATE \textbf{Step 2:} fix current optimal couplings $\Pi_l^i$ and update weight function $W_l$ using formula (\ref{eq: update formula for W_l});
		\UNTIL{$W_l$ and the $\Pi_l^i$ converge;}
		\STATE Obtain measure $\gamma_l$ based on $\gamma_{l-1}$ and $W_l$;\\
		\STATE Obtain optimal couplings $(\Pi_l^i)^* \in \Gamma(\gamma_l, \gamma_l^i)$ for $i=1, \cdots, n$.
		\ENDFOR
		\ENSURE 
		The new NN $N^{\text{new}}$ as specified by the measures $\gamma_1, \dots, \gamma_{m}$.
	\end{algorithmic}
\end{algorithm}

{ \begin{remark}[Total computational complexity for WB fusion]
	Without loss of generality, we assume that all the layers in each of the models has $M$ number of nodes. In practice, we set the maximum number of iterations of Steps 1 and 2 as $T$ (in the numerical experiments, we observe convergence within $T=10$).  From Proposition 1 in \cite{peyre2016gromov}, one can compute $\mathcal{L}(W_l, W_l^i) \otimes (\Pi_{l-1}^i)^*$ in $O(M^3)$ operations, and by using Sinkhorn algorithm \cite{cuturi2013sinkhorn}, the time complexity of solving an OT problem is roughly $O(M^2)$. Therefore, minimization with respect to $\{\Pi_l^i\}_i$ needs $O(nM^3 + nM^2)$ operations. On the other hand, minimization w.r.t. $W_l$ reduces to a simple matrix multiplication, which can be computed in $O(M^3)$ operations. Therefore, the complexity of updating $\{\Pi_l^i\}_i$ and $W_l$ for one round of Steps 1 and 2 is $O(nM^3)$. Then the total computational complexity of WB fusion algorithm is $O(nmM^3)$, where $m$ is the number of layers for the input models. Note that for modern neural networks, the number of nodes on each layer is not large, i.e., $M$ is relatively small. Therefore, our algorithm is quite fast especially when compared to training a NN from scratch.
\end{remark} 
}

\subsection{Extension to convolutional neural networks}\label{subsec: extension to CNNs}
\begin{figure}[htb]
	\centering
	\includegraphics[width=0.75\textwidth]{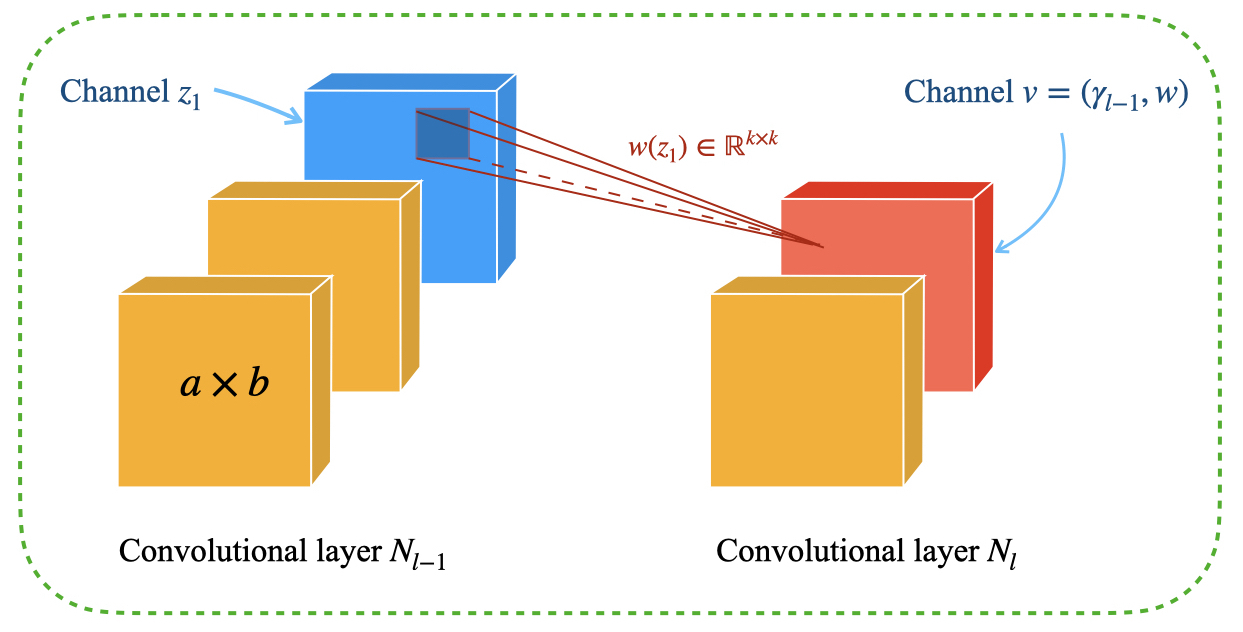}
	\caption{A channel $v$ in the convolutional layer $N_l$ is interpreted as a node in our framework, with weight function $w$ mapping previous layer elements (channels) to $\R^{k\times k}$.}
	\label{fig: CNN example}
\end{figure}

For CNNs, we substitute ``nodes'' for ``channels'' in the discussion in Sections \ref{sec: measure on neural networks} and \ref{sec: cost function}, and instead of defining probability measures and functions on sets of nodes, we define them on sets of channels . To be more concrete, let us consider a convolutional layer $N_l$ with weights having dimensions $ \mathbb{R}^{C^{\text{in}} \times k \times k \times C^{\text{out}}}$, where $C^{\text{in}}$, $C^{\text{out}}$ are the number of input and output channels and $k\times k$ is the size of filters. Then, in formula (\ref{def: definition of measure on layers}), $v$ must be interpreted as a channel on layer $N_l$, and its corresponding weight $w$ must be interpreted as a function mapping $N_{l-1}$ into $\mathbb{R}^{k \times k}$ (illustrated in Figure \ref{fig: CNN example}). Since now $w(z_q)$ is a $\mathbb{R}^{k \times k}$ matrix, we can redefine the cost function $d_{W}$ in (\ref{eqn: d_W}) using the Frobenius norm.

\subsection{Extension to ResNets}\label{subsec: extension to ResNets}
%For common fully connected or convolutional neural networks, layer $N_l^i$ is supported on the previous layer $N_{l-1}^i$. 

%\begin{wrapfigure}{R}{0.5\textwidth}
%	\vspace{-12pt}
%	\centering
%	\includegraphics[width=0.5\textwidth]{pics/resnet.png}
%	\caption{A building block of ResNet. Due to skip connection, 
%		outputs $X_1^i$ and $X_3^i$ of layers $N^i_1$ and $N^i_3$ gets added up
%		before feeding into layer $N^i_4$.}
%	\label{fig: ResNet's building block}
%	\vspace{-25pt}
%\end{wrapfigure}

\begin{figure}[htb]
	\centering
	\includegraphics[width=0.75\textwidth]{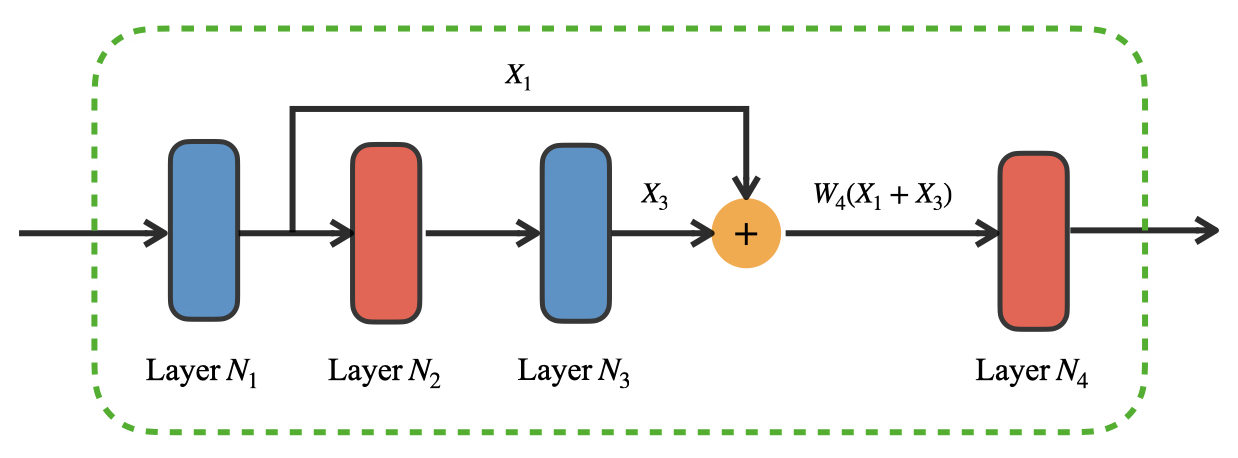}
	\caption{A building block of ResNet. Due to skip connection, outputs $X_1$ and $X_3$ of layers $N_1$ and $N_3$ are added up before feeding them to layer $N_4$.}
	\label{fig: ResNet's building block}
\end{figure}

{ResNet models are neural networks that contain skip connections
	 that ``jump'' over some layers to avoid the problem of vanishing gradients. Figure \ref{fig: ResNet's building block} shows a typical building block 
 	of a ResNet model. 
 	For a layer $N_l$ we let $\mathcal{C}_l=\{N_{\alpha}\}_{\alpha \in \Lambda}$ be the collection of all the previous layers that connect with it. Since the outputs of layers in $\mathcal{C}_l$ are added up before 
 	feeding them into layer $N_l$,  skip connections constrain the layers 
 	in $\mathcal{C}_l$ to share the same coupling, i.e $\Pi_\alpha=\Pi^*\; \forall \alpha$. This constraint allows us to use any of the previous layers in $\mathcal{C}_l$
	as the support for $N_l$ since they share the same coupling $\Pi^*$ 
	and hence produce the same cost to be used in optimization \eqref{eq: problem after plugging in cost function d_W}. Our convention is to find a coupling for the earliest layer 
	in $\mathcal{C}_l$ first and then use the same coupling when we fuse the other layers in $\mathcal{C}_l$ .
}

%\begin{figure}[htbp]
%	%\vspace{-12pt}
%	\centering
%	\includegraphics[width=0.6\textwidth]{pics/resnet.png}
%	\caption{A building block of ResNet. Due to skip connection, 
%		outputs $X_1^i$ and $X_3^i$ of layers $N^i_1$ and $N^i_3$ gets added up
%		before feeding into layer $N^i_4$.}
%	\label{fig: ResNet's building block}
%	%\vspace{-25pt}
%\end{figure}

%\begin{figure}[htbp]
%	\centering
%	\begin{subfigure}[b]{0.49\textwidth}
%		\centering
%		\includegraphics[width=\textwidth]{pics/cnn1.jpeg}
%		\caption{CNN.}
%		\label{fig: CNN example}
%	\end{subfigure}
%    \hfill
%    \begin{subfigure}[b]{0.49\textwidth}
%    	\centering
%    	\includegraphics[width=\textwidth]{pics/resnet1.jpeg}
%    	\caption{ResNet.}
%    	\label{fig: ResNet's building block}
%    \end{subfigure}
%    \caption{\textbf{Left:} A channel $v^i_{l,1}$ in the convolutional layer $N_l^i$ is interpreted as a node in our framework, with weight function $W^i_{l,1}$ mapping previous layer elements (channels) to $\R^{k\times k}$. \textbf{Right:} A building block of ResNet. Due to skip connection, outputs $X_1^i$ and $X_3^i$ of layers $N^i_1$ and $N^i_3$ gets added up before feeding into layer $N^i_4$.}
%\end{figure}

\section{Comparison between WB framework and OT framework}
{ In this section, we explain in detail how our Wasserstein barycenter-based method is different from OT fusion proposed in \cite{singh2019model}. Assume we have $n$ pre-trained models $N^1, \dots, N^n$ to aggregate. As before, denote the weights of the $l$-th layer in neural network $N^i$ and target model $N^{\text{new}}$ as $W_l^i$ and $W_l$ respectively. When doing the fusion on the $l$-th layer, OT fusion algorithm first aligns the incoming edge for the current layer $l$ by post-multiplying with the previous layer transport matrix $(\Pi_{l-1}^i)^*$ and normalizing via the inverse of the probability measure over the $l$-th layer in the target model, i.e.,
\begin{equation}\label{eqn: aligning incoming edge weights in OT fusion}
\widehat{W}_{l}^i \leftarrow k_{l-1}\frac{1}{\mathds{1}_{k_l-1}}W_{l}^i (\Pi_{l-1}^i)^{*T}
\end{equation}
Then the neurons in layer $l$ of the pre-trained models $N_l^i$ are aligned with respect to the target model $N^{\text{new}}$ by considering optimal transport problems with cost matrices:
\begin{equation}\label{eqn: cost function defined in OT fusion}
C_{l, \text{OT}}^i := \big[\|w_j - \widehat{w}_ g^i\|_2^2 \big]_{j, g},
\end{equation}
where $w_j$ is the $j$-th row vector of weight matrix $W_l$ and $\widehat{w}_g^i$ is the $g$-th row vector of $\widehat{W}_{l}^i$. On the other hand, the cost function in our WB framework reads:
\begin{equation}\label{eqn: cost functions in our WB-based fusion}
    C_{l, \text{WB}}^i := \big[c_l(v_{j}, v_{g}^i)\big]_{j, g} = \big[d_W(w_{j}, w_{g}^i) \big]_{j,g} = \big[(w_{jq} - w_{gs}^i)^2\big]_{j,g,q,s} \otimes (\Pi_{l-1}^i)^*
\end{equation}
We can see that the cost functions used in our algorithm \eqref{eqn: cost functions in our WB-based fusion} are different from the ones in OT fusion \eqref{eqn: cost function defined in OT fusion} .
% And the definition of our cost function is motivated from the idea of $TL^p$ distance as mentioned in Section \ref{sec: explanation on cost function}. In particular, since we interpret $W_{l,j}$ and $W_{l, h}^i$ as functions supported on measures $\gamma_{l-1}$ and $\gamma_{l-1}^i$ respectively, it is naturally to define the distance between $W_{l,j}$ and $W_{l, h}^i$ based on the optimal couplings $(\Pi_{l-1}^i)^*$ between their measure supports. 
We highlight that our cost function \eqref{eqn: cost functions in our WB-based fusion} was derived from a first principle, in this case a bona fide Wasserstein barycenter problem. In contrast, the aligning step \eqref{eqn: aligning incoming edge weights in OT fusion} in OT fusion is introduced without a similar derivation.
%, thus its connection to Wasserstein barycenters is unclear. 

Since our target problem is from the beginning a minimization problem, it is clear that one should iterate the proposed steps until convergence, rather than stopping after one iteration as in OT fusion. As illustrated in the numerical experiments (Section \ref{sec: WB hetero}, \ref{sec: WB homo}), iterating until convergence is important.}

\section{Details on GWB fusion}\label{sec: GWB Fusion}
\begin{figure}[htb]
	\centering
	\includegraphics[width=0.75\textwidth]{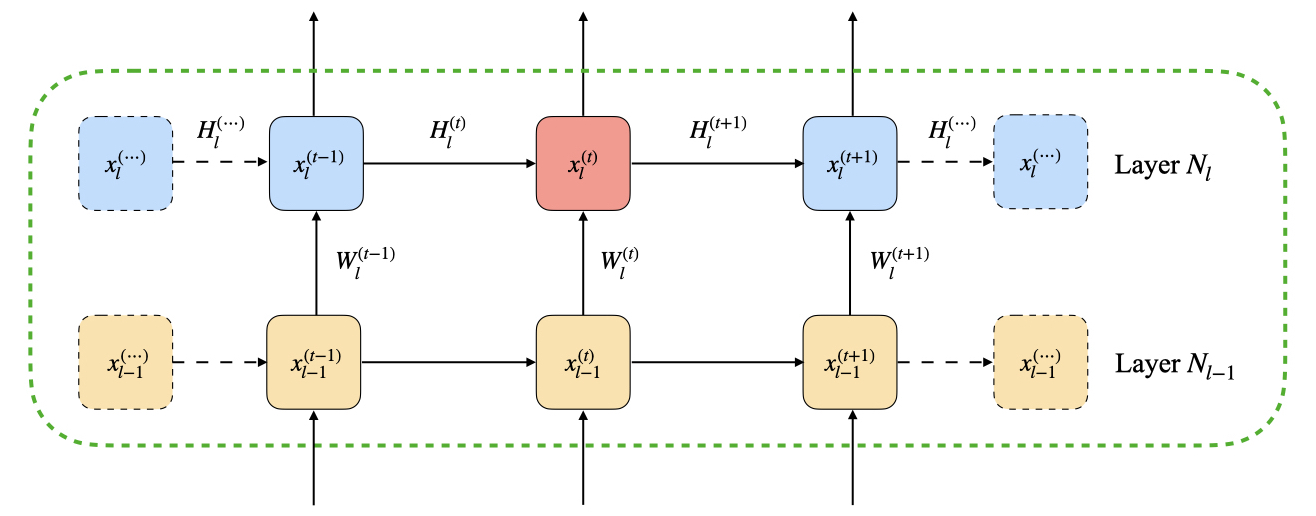}
	\caption{A building block of unfolded RNN motivating problem \eqref{eq: GWBP for rnn fusion}. $W_l^{(t)}$ and $H_l^{(t)}$ are collections of input-to-hidden and hidden-to-hidden weights functions at time step $t$ respectively (for actual RNNs, $W_l$ and $H_l$ don't depend on $t$). $x_{l}^{(t)}$ denotes the output of layer $N_l$ at time step $t$; $t$ changes horizontally and indexes the unfolded units.}
	\label{fig: RNN's building block}
\end{figure}  

% \begin{wrapfigure}{R}{0.5\textwidth}
%     \vspace{-15pt}
% 	\centering
% 	\includegraphics[width=0.5\textwidth]{pics/rnn.png}
% 	\caption{A building block of unfolded RNN motivating problem \eqref{eq: GWBP for rnn fusion}. $W_l^{(t)}$ and $H_l^{(t)}$ are collections of input-to-hidden and hidden-to-hidden weights functions at time step $t$ respectively (for actual RNNs, $W_l$ and $H_l$ don't depend on $t$). $x_{l}^{(t)}$ denotes the output of layer $N_l$ at time step $t$; $t$, which changes horizontally, indexes unfolded units.}
% 	\label{fig: RNN's building block}
% 	\vspace{-12pt}
% \end{wrapfigure}
{Analogous to the fusion of FC networks, 
we can think of RNN fusion on the $l$-th layer following the ``layerwise'' fashion with respect to the ``time step''. Specifically, consider the unfolded RNN shown in Figure \ref{fig: RNN's building block} and the fusion for layer $l$ at time step $t$.  We assume that the fusion before time step $t$ has finished.} Denote $v^{(t)}_j= [(\gamma_{l-1}, w^{(t)}_j); (\gamma_l^{(t-1)}, h^{(t)}_j)]$ and $v_g^i= [(\gamma_{l-1}^i, w_g^i); (\gamma_l^i, h_g^i)]$ as the $j$-th node in layer $N_l$ at time $t$ and the $g$-th node in layer $N_l^i$ respectively. Then the cost function between nodes $v_j^{(t)}$ and $v_g^i$ at time $t$ could be defined as
\begin{equation}\label{eq: cost function of RNN}
	c_l(v^{(t)}_j, v_g^i) := \underbrace{d_{\mu}(\gamma_{l-1}, \gamma_{l-1}^i) + d_{W}(w^{(t)}_j, w_g^i)}_\text{ $TL^p$ cost for input-to-hidden weights}
	+ \,\underbrace{d_{\mu}(\gamma_{l}^{(t-1)}, \gamma_{l}^i) + d_{H}(h^{(t)}_j, h_g^i)}_\text{$TL^p$ cost for hidden-to-hidden weights},
\end{equation}
where $d_{\mu}(\cdot, \cdot)$, $d_{W}(\cdot, \cdot)$ are defined the same as \eqref{eq: dMu} and \eqref{eqn: d_W} respectively. Denote 
\begin{equation}
\Pi_l^{i, (t-1)} := \argmin_{\Pi_l^i} d_{ \mu}(\gamma_{l}^{(t-1)}, \gamma_{l}^i) = \argmin_{\Pi_l^i} \langle C_l^{i, (t-1)}, \Pi_{l}^i \rangle,
\end{equation}
where $C_{l}^{i, (t-1)} := [c_{l}(v_j^{(t-1)}, v_g^i)]_{j, g}$ is the cost matrix for $l$-th layer at time $t-1$.
We use $H_l^{(t)} := (h_1^{(t)}, h_2^{(t)}, \dots, h_{k_l}^{(t)})^T$ to denote the hidden-to-hidden weight function matrix of layer $N_l^{\text{new}}$ at time step $t$. Since $H_l^{(t)}$ is supported on the $l$-th layer (at time step $t-1$), we define
\begin{equation}
	d_{H}(H_{l}^{(t)}, H_{l}^i) := \mathcal{L}(H_l^{(t)}, H_l^i)  \otimes \Pi_{l}^{i, (t-1)},
\end{equation}
where $\mathcal{L}(H_l^{(t)}, H_l^i) := \big[\big(h_{jq}^{(t)} - h_{gs}^i\big)^2\big]_{j,g,q,s}$ is a 4-way tensor. Now consider the WBP (\ref{eq: formal WB on l-th layer}) again (plugging-in the cost function \eqref{eq: cost function of RNN}). 
Note that  $\gamma_{l}^{(t-1)}$ is given, then $d_{\mu}(\gamma_{l}^{(t-1)}, \gamma_{l}^i)$ becomes a constant and does not involve any optimization variable in (\ref{eq: formal WB on l-th layer}) like $d_{\mu}(\gamma_{l-1}, \gamma_{l-1}^i)$. Therefore, the barycenter problem on the $l$-th layers at time step $t$ is to minimize the objective function
\begin{equation}
		B(W_l^{
		(t)}, H_l^{(t)}; \{\Pi_l^{i, (t)}\}_i) := 
	    \frac{1}{n} \sum_{i=1}^{n} \langle \mathcal{L}(W_l^{(t)}, W_l^i) \otimes (\Pi_{l-1}^i)^* + \alpha_H\mathcal{L}(H_l^{(t)}, H_l^i) \otimes \Pi_{l}^{i, (t-1)}, \Pi_{l}^{i, (t)}\, \rangle,
\end{equation}
with respect to $W_l^{(t)}, H_l^{(t)}$ and $\{\Pi_l^{i, (t)}\}_i$. However, since RNNs presuppose that all weights should be the same for all unfolded units (i.e. for all $t$ along a fixed $l$), it is natural to consider invariants (i.e. independent of $t$) of the ensemble of problems indexed by $t$. One such invariant can be obtained by formally taking the limit $t \rightarrow \infty$ in the above problem: this gives rise to the following problem
\begin{equation}\label{eq: GWBP for rnn fusion (appendix)}
		\min_{W_l, H_l, \{\Pi_l^i\}_i}  B(W_l, H_l; \{\Pi_l^i\}_i) := 
		\frac{1}{n} \sum_{i=1}^{n} \langle \mathcal{L}(W_l, W_l^i) \otimes (\Pi_{l-1}^i)^* + \alpha_H\mathcal{L}(H_l, H_l^i) \otimes \Pi_{l}^i, \Pi_{l}^i\, \rangle
\end{equation}

\subsection{GWB fusion algorithm}
Notice that \eqref{eq: GWBP for rnn fusion} is a GW-like barycenter problem. Therefore, following the algorithm proposed in \cite{peyre2016gromov}, we solve it using a  block coordinate relaxation, i.e., alternatively minimizing with respect to the couplings $(\Pi_{l}^i)_i$ and weight functions $W_l$ and $H_l$.

\textbf{Minimization with respect to $\{\Pi_{l}^i\}_i$.}  The optimization (\ref{eq: GWBP for rnn fusion}) over $(\Pi_{l}^i)_i$ alone decouples as $n$ independent GW-like optimization problems. For $i = 1, \cdots, n$
\begin{equation}
	\min_{\Pi_{l}^i \in \Gamma(\gamma_l, \gamma_l^i)} \, \langle \mathcal{L}(W_l, W_{l}^i) \otimes (\Pi_{l-1}^i)^*+ \mathcal{L}(H_l, H_{l}^i) \otimes \Pi_{l}^i, \, \Pi_l^i \rangle.
\end{equation}
As proposed in \cite{peyre2016gromov}, a stationary point of this optimization problem can be reached following the iterations:
\begin{equation}\label{eq: updating with respect to Pi_l^i}
	\Pi_{l}^i \leftarrow \mathcal{T}\big(\mathcal{L}(W_l, W_{l}^i) \otimes (\Pi_{l-1}^i)^*+ \mathcal{L}(H_l, H_{l}^i) \otimes \Pi_{l}^i, \gamma_l, \gamma_l^i\big).
\end{equation}

\textbf{Minimization with respect to $W_l$ and $H_l$.} For given $\{\Pi_{l}^i\}_i$, the minimization with respect to $W_l$ and $H_l$ reads
\begin{equation}
	\min_{W_l, H_l} \, \frac{1}{n}\sum_{i=1}^{n}\langle \mathcal{L}(W_l, W_{l}^i) \otimes (\Pi_{l-1}^i)^*+ \mathcal{L}(H_l, H_{l}^i) \otimes \Pi_{l}^i, \, \Pi_l^i \rangle.
\end{equation}
By first-order optimality conditions, we have the update formulas,
\begin{align}
	W_l &\leftarrow k_l k_{l-1}\frac{1}{\mathds{1}_{k_{l-1}}\mathds{1}_{k_l}^T} \frac{1}{n}\sum_{i=1}^{n} \Pi_{l}^i W_l^i (\Pi_{l-1}^i)^{*T},
	\label{eq: updating formula for W_l}\\ 
	H_l &\leftarrow k_l^2\frac{1}{\mathds{1}_{k_l}\mathds{1}_{k_l}^T} \frac{1}{n}\sum_{i=1}^{n} \Pi_{l}^i H_l^i (\Pi_{l}^i)^T \label{eq: updating formula for H_l}
\end{align}

The GWB fusion algorithm is summarized in Algorithm \ref{alg: rnn fusion}.

\begin{algorithm}[H]
	%\flushleft
	\caption{GWB Fusion}
	\label{alg: rnn fusion}
\begin{algorithmic}
		\REQUIRE
		Neural networks $N^1, \cdots, N^n$ ($m$ layers);\\
		Number of nodes $k_l$ for layer $N_{l}^{\text{new}}$,  for $l =2 , \cdots, m$;\\
		Initialized weight functions $\{W_l\}_{l=2}^{n}$ and $\{H_l\}_{l=2}^{n}$;\\
		Set $(\Pi_1^i)^*$ to be the identity matrix in $\mathbb{R}^{k_1}$ for all $i=1, \dots, n$. Set also $\gamma_1= \gamma_1^1$;
		\FOR{$l = 2, \cdots, m$}
		%		\STATE Choose $\mu_l = \gamma_{l-1}$ and define $d_W^2$ based on couplings $(\Pi_{l-1}^i)^*$'s;\\
		\REPEAT 
		\FOR{$i=1, \cdots, n$} 
		\STATE Initialize $\Pi_{l}^i$;
		\REPEAT
		\STATE Compute $C_l^i := \mathcal{L}(W_l, W_{l}^i) \otimes (\Pi_{l-1}^i)^*+ \mathcal{L}(H_l, H_{l}^i) \otimes \Pi_{l}^i$;
		\STATE Updating $\Pi_l^i$ by solving entropy-regularized optimal transport problem \eqref{eq: updating with respect to Pi_l^i};
% 		\STATE Initialize $a \leftarrow \mathds{1}_{k_l}$, set $K \leftarrow e^{-C_l^i/\varepsilon}$;
% 		\REPEAT %{\blue\Comment{Sinkhorn iterations (\ref{eq: sinkhorn iterations}) to compute $\mathcal{T}(C_l^i, \gamma_l, \gamma_l^i)$}}
% 		\STATE $a \leftarrow \frac{1}{k_l}\frac{\mathds{1}_{k_l}}{Kb}, b \leftarrow \frac{1}{k_l^i} \frac{\mathds{1}_{k_l^i}}{K^T a}$;
% 		\UNTIL{$a$ and $b $ converge;}
% 		\STATE Update $\Pi_{l}^i \leftarrow \text{diag}(a)K\text{diag}(b)$;
		\UNTIL{$\Pi_{l}^i$ converges;}
		\ENDFOR
		\STATE Updating $W_l$ and $H_l$ using (\ref{eq: updating formula for W_l}) and (\ref{eq: updating formula for H_l}); %{\blue \Comment{minimize over $W_l$ and $H_l$}}
		\UNTIL{$W_l$, $H_l$ and $\{\Pi_{l}^i\}_i$ converge;}
		\STATE Obtain measure $\gamma_l$ based on $\gamma_{l-1}$, $W_l$ and $H_l$;
		\STATE Obtain optimal couplings $(\Pi_l^i)^* \in \Gamma(\gamma_l, \gamma_l^i)$ for $i=1, \cdots, n$;
		\ENDFOR
		\ENSURE
		The new NN $N^{\text{new}}$ as specified by the measures $\gamma_1, \dots, \gamma_{m}$.
\end{algorithmic}
\end{algorithm}

{ \begin{remark}[Total computational complexity for GWB fusion]
Similar to the analysis of WB fusion, we assume each layer has $M$ number of neurons and we set the maximum number of times we run the outer repeat loop in Algorithm \ref{alg: rnn fusion} to be $T$. Also, we let $\tilde{T}$ be maximum number of times we run the inner repeat loop in Algorithm \ref{alg: rnn fusion} (in our numerical experiments, we observe convergence within $\tilde T =10$). From Proposition 1 in \cite{peyre2016gromov}, one can compute the cost matrix $C_l^i = \mathcal{L}(W_l, W_{l}^i) \otimes (\Pi_{l-1}^i)^*+ \mathcal{L}(H_l, H_{l}^i) \otimes \Pi_{l}^i$ in $O(M^3)$ operations. The time complexity of solving entropy-regularized OT problem \eqref{eq: updating with respect to Pi_l^i} using Sinkhorn algorithm is $O(M^2)$. Therefore, minimizing with respect to $\{\Pi_l^i\}_i$ needs $O(n\tilde{T}M^3 + n\tilde{T}M^2)$ operations. On the other hand, minimizing w.r.t $W_l$ and $H_l$ can be computed in $O(M^3)$ operations. Therefore, the complexity of updating $\{\Pi_l^i\}_i$, $W_l$ and $H_l$ for one iteration is $O(nM^3)$. Then, the total computational complexity of GWB fusion algorithm is $O(nmM^3)$.
\end{remark}
}

\subsection{Extension to LSTM}
{
Our GWB framework for RNN models can be easily extended to the case of LSTMs.
LSTMs are more sophisticated than RNNs. 
For the purpose of fusion the major difference is 
that LSTMs have $4$ cell states (hidden states) in comparison to RNNs which have one hidden state.
We consider each cell of LSTM as an individual unit during the fusion process, 
%Fusion of LSTM can be easily extended from the basic RNN case. In particular, we regard each cell state of LSTM as individual units during the fusion, 
i.e., treating each of the cells as a basic RNN unit. 
However, since these cells share the same input from the previous layer, 
the alignments of cells should also be the same, 
and thus we add the constraint that 
the coupling matrices corresponding to each cell should be the same. 
To reflect this constraint, 
the cost function for each hidden layer in LSTM 
is updated to be the sum of cost functions defined for each cell unit.}

\section{Experiment Details}
\label{sec: details}

In this section, we provide more details on model training 
and related hyperparameters for our experiments.

% \subsection{Evaluation methodology}
% We evaluate our framework in terms of task accuracy on an unseen test set. 
% The fully connected networks are evaluated on MNIST dataset 
% which consists of images of digits and its labels, while
% the deep convolutional neural networks, $\vgg$ and $\resnet$, are evaluated on 
% standard image classification dataset like CIFAR10 
% which consists of images from
% 10 different classes - airplanes, cars, birds, cats, deer, dogs, frogs, horses, ships, and trucks.

\subsection{Model training}
\textbf{$\mlp$ training details.} For the fusion experiments,
$\mlp$, $\mlps$ and $\mlpl$ models are trained using SGD optimizer at a constant
learning rate of $0.05$ and momentum of $0.5$ for $20$ epochs. 
In the fusion experiments for heterogeneous data distributions, we follow \cite{singh2019model}
and training the $\mlps$ models using SGD optimizer with learning rate of $0.01$, momentum of $0.5$
for $10$ epochs.

\textbf{$\vgg$ training details}. We follow \cite{singh2019model},
and train the $\vgg$ model using SGD optimizer for $300$ epochs. 
The initial learning rate of $0.05$ decays after every $30$ epochs by a 
multiplicative factor of $0.5$. We use SGD with momentum of $0.9$ and
weight decay of $0.0005$. The batch size used in training is $128$.
The model with best validation accuracy is selected for fusion. 

\textbf{$\resnet$ training details.} The models are trained using SGD optimizer 
for $300$ epochs with an initial learning rate of $0.1$ which gets decayed by a multiplicative
factor of $0.1$ at epoch $150$. We use momentum of $0.9$ and weight decay of $0.0001$.

We skip the batch normalization layer in $\resnet$ models
for the current work. However, our framework can be extended to handle batch normalization
parameters by appropriately including them as a part of the node and using the coupling
associated with the previous layers. We leave this extension for the future work.

{\textbf{RNN training details.} 
For all our experiments the RNN model has one hidden layer of dimension $256$.
%We use 4 different datasets for the RNN experiments. 
In general, the RNN base models are trained using Adam \cite{kingma2014adam} optimizer. 
We use momentum of $0.9$ and weight decay of $1 \times 10^{-4}$. 
For MNIST dataset, the images of $28 \times 28$ dimensions are 
interpreted as 28 length sequences of vectors $\in \mathbb{R}^{28}$.
% and hence the RNN model has input dimension of $28$.
For the NLP tasks, we use pre-trained GloVe embeddings \cite{pennington2014glove} of dimension $100$.
The pre-trained embedding layer does not get updated during the training phase.
The dataset specific training details are as follows:
%In the following, we list the settings that vary for each dataset. 
i) MNIST: Models are trained for $20$ epoch with a constant learning rate of $0.001$ and 
a batch size of $64$; 
ii) SST-2: Models are trained for $20$ epochs with a constant learning rate of $0.001$ and batch size of $256$. 
The maximum sequence length for the dataset is set to $56$;  
iii) AGNEWS: Models are trained for $10$ epochs. 
The constant learning rate is $0.0001$ and the batch size used in training is $128$.
The maximum sequence length for each training data is set to $60$ 
for stable training of RNN;
and 
iv) DBpedia: Models are trained for $30$ epochs with a constant learning rate $0.0001$, batch size $256$ and
a maximum sequence length of $60$.
}

{\textbf{LSTM training details.} 
Similar to the RNN case, for all our experiments the LSTM model has one hidden layer of hidden dimension $256$. 
Since LSTMs have $4$ hidden states, this corresponds to a hidden layer of total $4\times 256$ dimensions. 
%we also use $4$ different datasets for LSTM training. 
In general, the LSTM base models are trained using Adam \cite{kingma2014adam} optimizer with a constant learning rate $0.001$. 
We use momentum of $0.9$ and weight decay of $1 \times 10^{-4}$. 
For the NLP tasks, we use pre-trained GloVe embeddings \cite{pennington2014glove} of dimension $50$
and the pre-trained embedding layer does not get updated during the training phase.
As for the RNNs, the images of $28 \times 28$ dimensions in MNIST dataset 
are interpreted as 28 length sequences of vectors $\in \mathbb{R}^{28}$.
The dataset specific training details are as follows:
%In the following, we list the settings that vary for each dataset: 
i) MNIST: Models are trained for $50$ epochs with batch size of $64$; 
ii) SST-2: Models are trained for $20$ epochs. The batch size used for training is $256$. 
The maximum sequence length is set to $56$; 
iii) AGNEWS: Models are trained for $10$ epochs with batch size $128$ and a maximum sequence length of $160$; 
and iv) DBpedia: Models are trained for $5$ epochs and the batch size used for training is $256$.
The maximum sequence length for this dataset is set to $100$.}

% \textbf{General.} The bias terms in linear layers of all the networks are set to zero in all the experiments.
% A straightforward modification of our setup can be used to handle the general case,
% which we leave for future work.

\subsection{Hyperparameter selection}
The hyperparameters are selected using a separate validation split. 
For solving the optimal transport problem in our proposed framework, we use Sinkhorn algorithm
\cite{cuturi2013sinkhorn} with regularization hyperparameters $0.01, 0.005, 0.001, 0.0005$. 
The hyperparameter $\alpha_H$ in \eqref{eq: GWBP for rnn fusion} is chosen in the interval $[1, 20]$. 
{Note that in our implementation, we use the variable $alpha\_h$ to capture $\alpha_H$
and set it between $[100, 2000]$ since we simplify the implementations and
don't normalize the coupling matrices $\Pi_1^i$ by the size of layers.}

\section{Additional experiments and results}

\subsection{Fusion of multiple model counts}
{\textbf{Setup:} 
In this experiment, we consider fusion of $2$, $4$ and $6$ base models into a target model of different architecture. 
We use $\mlp$ trained on MNIST dataset as the base models. 
The target model is set to be $\mlpl$ which has twice the width of the base models.
Since the base and target models have different architectures,
we initialize the target model randomly before starting fusion.

\textbf{Quantitative results:} Table \ref{table: WB fusion for multiple model counts} 
shows the results for fusion of multiple models. 
% The accuracy values are reported over $9$ random runs. 
We find that the WB framework performs much better than OT fusion algorithm 
(higher accuracy and low standard deviations).
These improvements in accuracy are especially higher
when the number of base models being fused is large. 
These experiments illustrate that in comparison to OT fusion 
WB fusion is much more robust to randomness in initialization.
%These experiments again illustrate the robustness of 
%WB framework in comparison to OT fusion algorithm.
}

\begin{table}[!htb]
	\caption{Performance comparison (Test accuracy $\pm$ standard deviation \%) of OT and WB frameworks for fusing multiple models into a different target architecture. For each case, the target model obtained by WB fusion has higher test accuracy and smaller standard deviation.}
	\label{table: WB fusion for multiple model counts}
	\vspace{4pt}
	\centering
	\renewcommand\arraystretch{1.5}
	\resizebox{0.8\textwidth}{!}{
		\begin{sc}
			\begin{tabular}{@{}cccc@{}}
				%\toprule
				& \multicolumn{3}{c}{MNIST/$\mlpl$}   \\ 
				\cmidrule{2-4} 
				&  \# of Models=2 & \# of Models=4 & \# of Models=6  \\
				\midrule
				Base Model Avg & $98.31 \pm 0.02$ &$98.31 \pm 0.02$ & $98.31 \pm 0.02$  \\
				\midrule[0.05pt]
				OT & $91.53 \pm 2.64$ & $52.10 \pm 8.35$ &  $43.16 \pm 6.17$ \\
				WB & \bm{$94.93 \pm 1.18$}& $\bm{89.03 \pm 4.05}$ & $\bm{86.40 \pm 4.29}$ \\
				\bottomrule
			\end{tabular}
		\end{sc}
	}
\end{table}

\subsection{Further visualizations under different network architectures and datasets}
\begin{figure}[htb]
	\centering
	\begin{subfigure}{.5\textwidth}%
    \includegraphics[width=1.0\linewidth]{pics/test_error_plane_VGG11_CIFAR10.pdf}
    \end{subfigure}%
	\begin{subfigure}{.5\textwidth}%
    \includegraphics[width=1.0\linewidth]{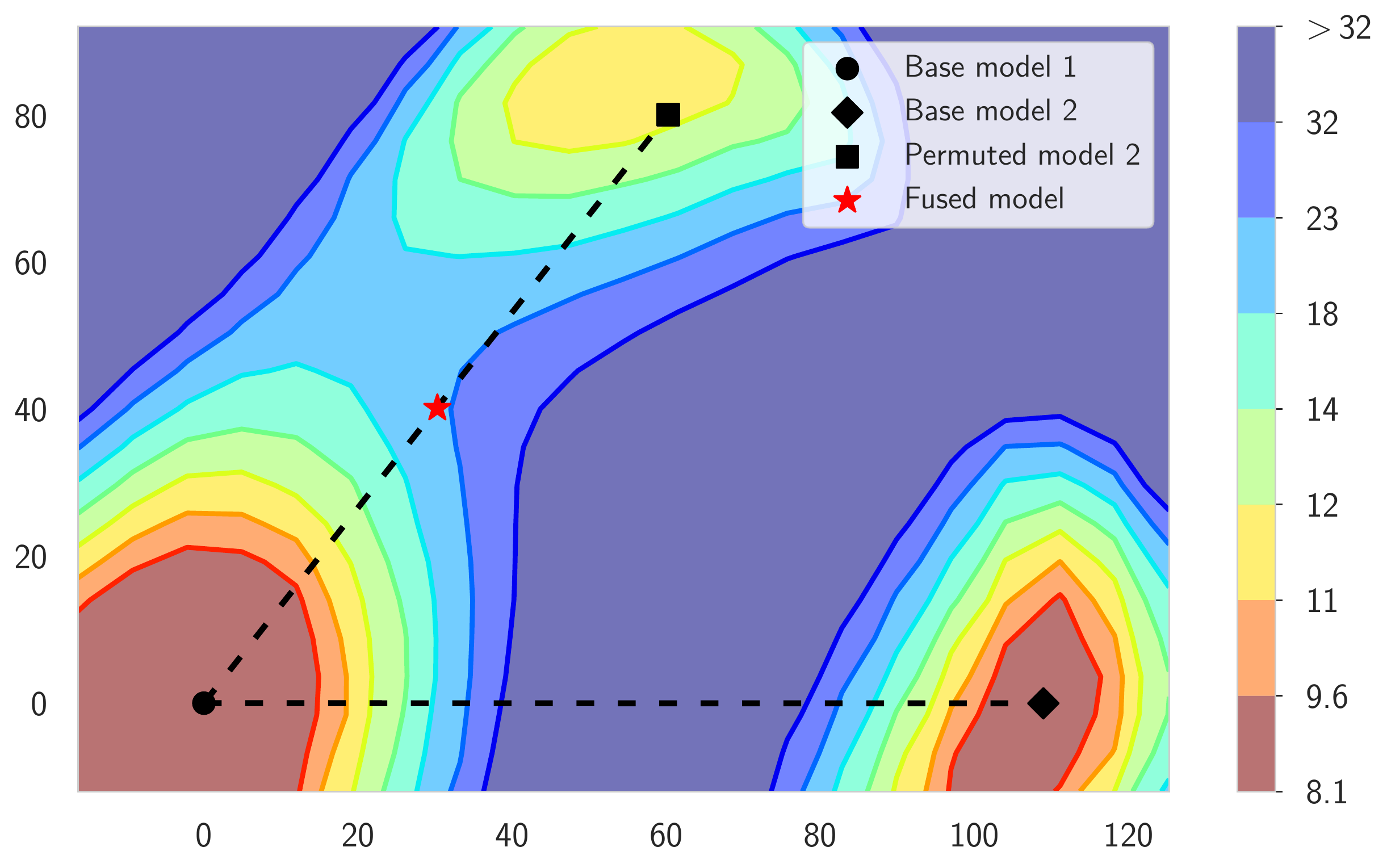}
    \end{subfigure}
    \caption{The test error surface of (\textbf{Left}) $\vgg$ trained on CIFAR10 dataset, (\textbf{Right}) $\resnet$ trained on CIFAR10 dataset.}
	\label{fig: visualization for CIFAR10 dataset}
\end{figure}

\begin{figure}[!htb]
    \centering
    \begin{subfigure}{.5\textwidth}%
    \includegraphics[width=1.0\linewidth]{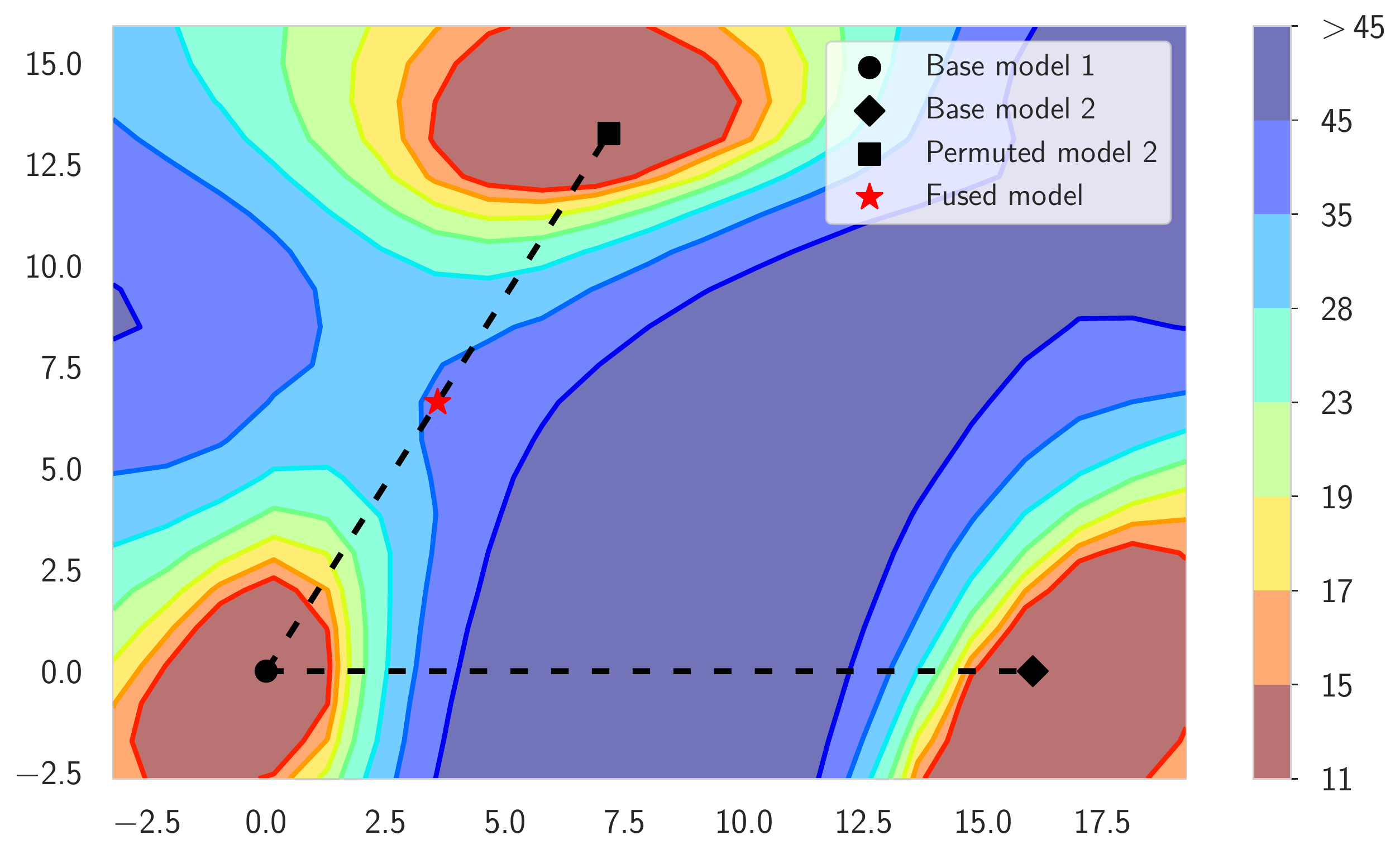}
    \end{subfigure}%
    \begin{subfigure}{.5\textwidth}
    \includegraphics[width=1.0\linewidth]{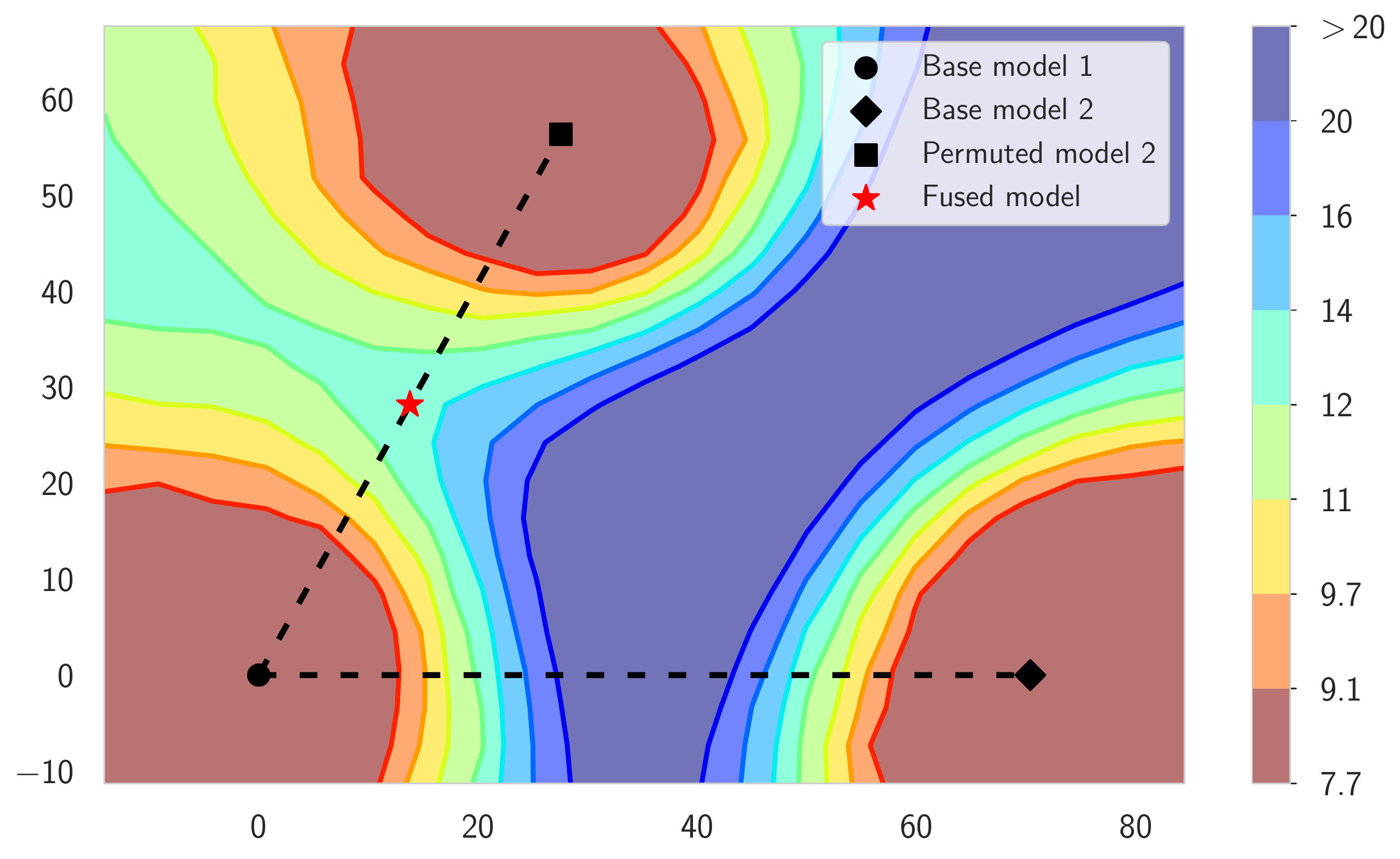}
    \end{subfigure}
\caption{The test error surface of (\textbf{Left}) RNN trained on AGNEWS dataset, (\textbf{Right}) LSTM trained on AGNEWS dataset.}
\label{fig: visualization for AGNEWS dataset}
\end{figure}

\begin{figure}[!htb]
    \centering
    \begin{subfigure}{.5\textwidth}
    \includegraphics[width=1.0\linewidth]{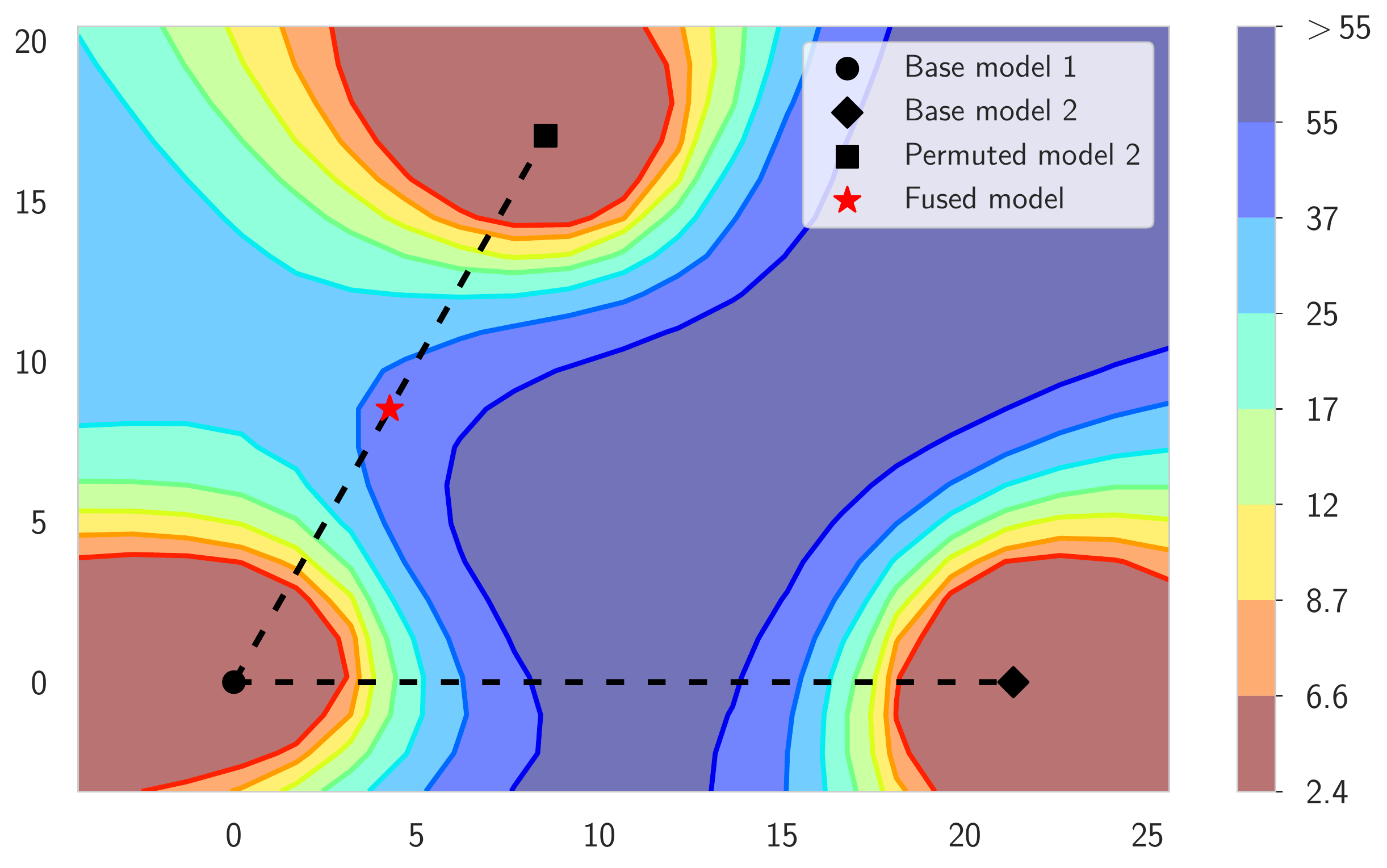}
    \end{subfigure}%
    \begin{subfigure}{.5\textwidth}
    \includegraphics[width=1.0\linewidth]{pics/test_error_plane_LSTM_DBpedia.pdf}
    \end{subfigure}
\caption{The test error surface of (\textbf{Left}) RNN trained on DBpedia dataset,  (\textbf{Right}) LSTM trained on DBpedia dataset.}
\label{fig: visualization for DBpedia dataset}
\end{figure}

{In this section, we include more visualizations of the models produced by our algorithms when fusing different network architectures trained over different datasets \footnote{We use the visualization method proposed in \cite{garipov2018loss}; their code is available at \url{https://github.com/timgaripov/dnn-mode-connectivity}}. We find that in each setting, the basins of the permuted model 2 and base model 1 lie close to each other and are separated by a relatively low energy barrier, especially when compared to the energy barriers between the basins of model 1 and model 2. However, we also observe that, for cases like $\resnet$ trained on CIFAR10 dataset (Figure \ref{fig: visualization for CIFAR10 dataset} (right)) or RNN trained on AGNEWS and DBpedia datasets (Figure \ref{fig: visualization for AGNEWS dataset} (left) and Figure \ref{fig: visualization for DBpedia dataset} (left)), the energy barrier between the basins of model 1 and permuted model 2 is not as low as for our other experiments. This might be attributed to specific properties of the model types (deep network architectures) and datasets. We leave further exploration of this as an important future work. }

\begin{figure}[htb]
    \centering
    \begin{subfigure}{.5\textwidth}%
    \includegraphics[width=1.0\linewidth]{pics/test_error_plane_FCNN_MNIST.pdf}
    \end{subfigure}
    \begin{subfigure}{.5\textwidth}%
    \includegraphics[width=1.0\linewidth]{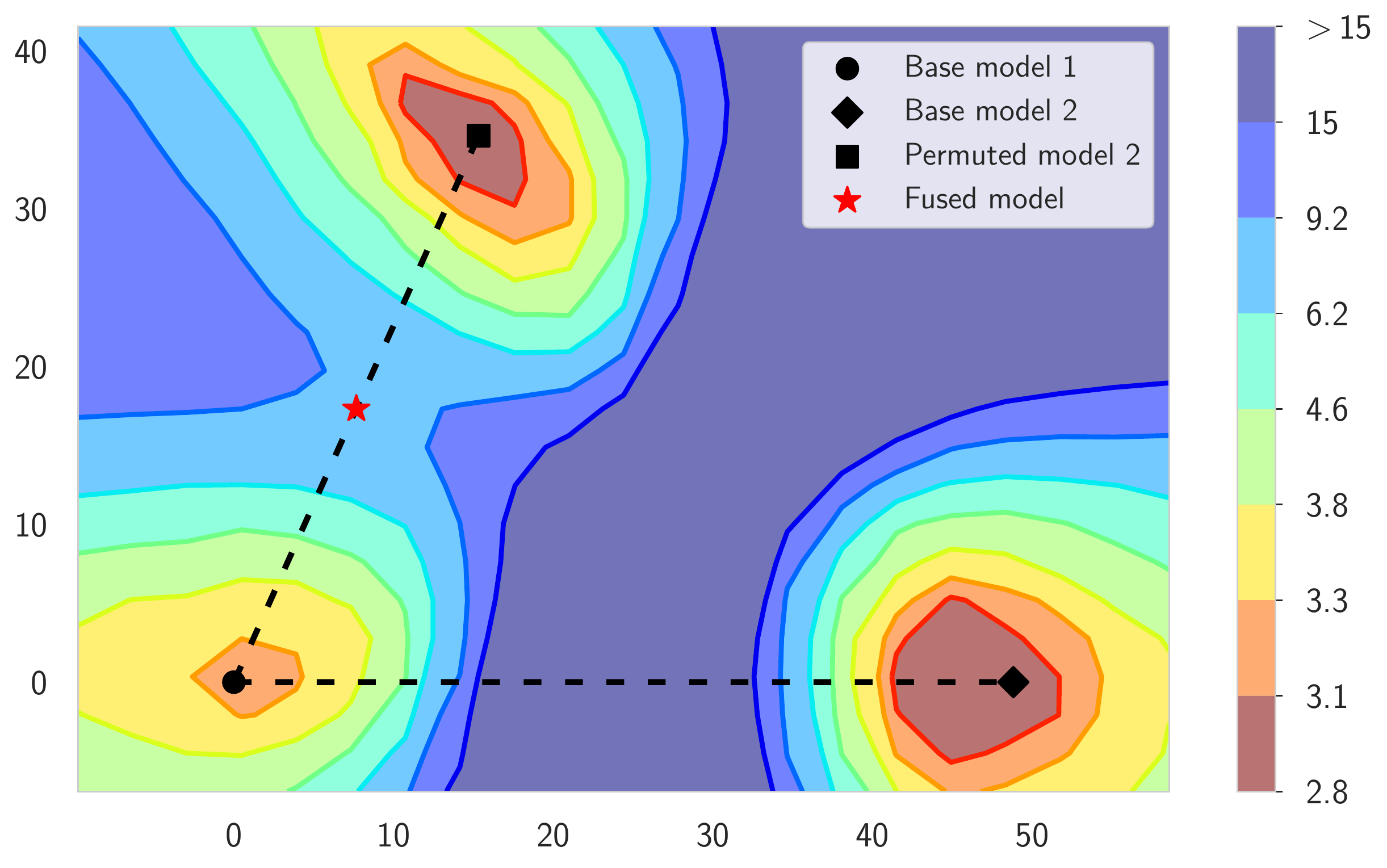}
    \end{subfigure}%
    \begin{subfigure}{.5\textwidth}
    \includegraphics[width=1.0\linewidth]{pics/test_error_plane_LSTM_MNIST.pdf}
    \end{subfigure}
\caption{The test error surface of (\textbf{Top}) $\mlp$ trained on MNIST dataset,  (\textbf{Bottom Left}) RNN trained on MNIST dataset, (\textbf{Bottom Right}) LSTM trained on MNIST dataset.}
\label{fig: visualization for MNIST dataset}
\end{figure}

\begin{figure}[!htb]
    \centering
    \begin{subfigure}{.5\textwidth}%
    \includegraphics[width=1.0\linewidth]{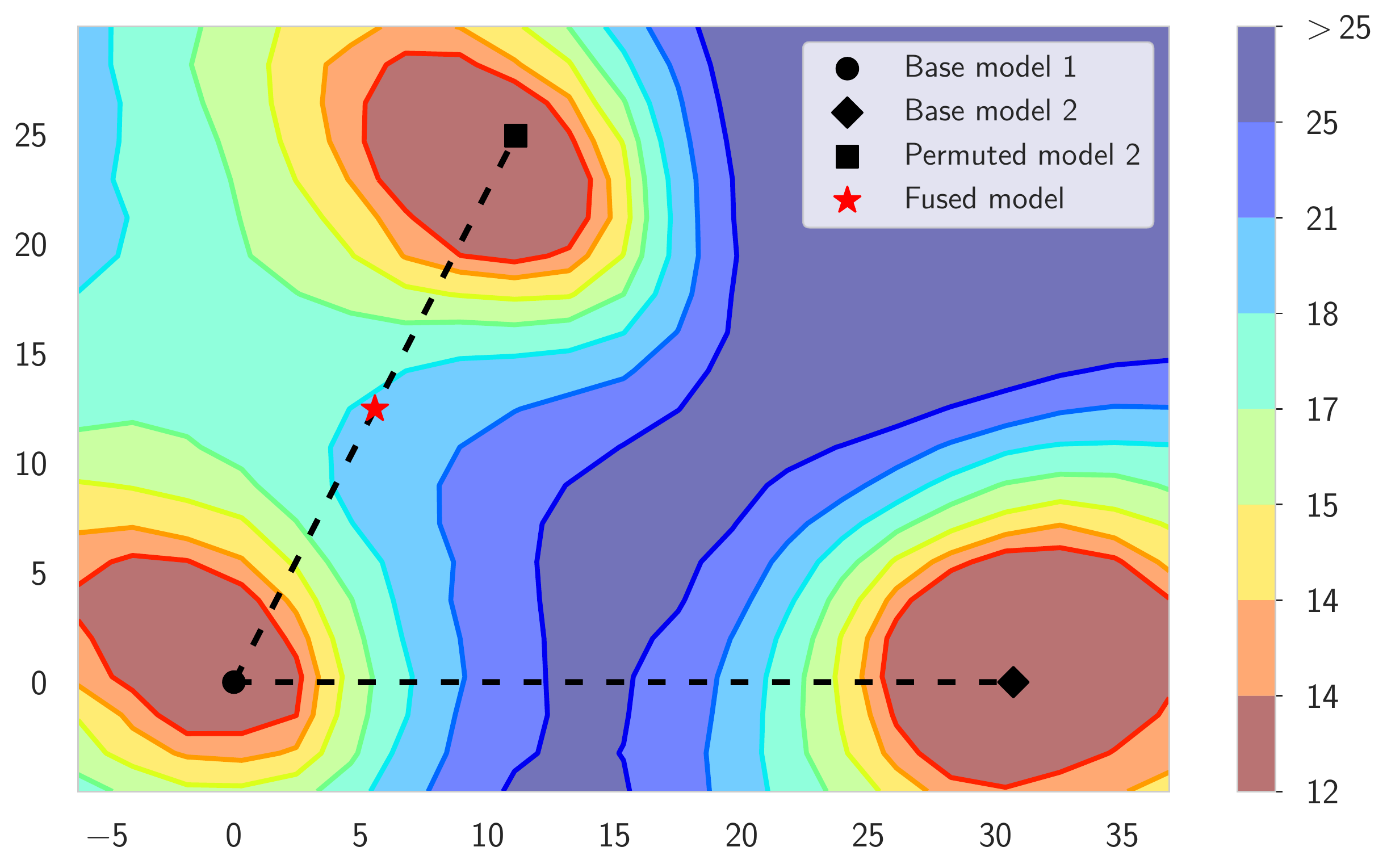}
    \end{subfigure}%
    \begin{subfigure}{.5\textwidth}
    \includegraphics[width=1.0\linewidth]{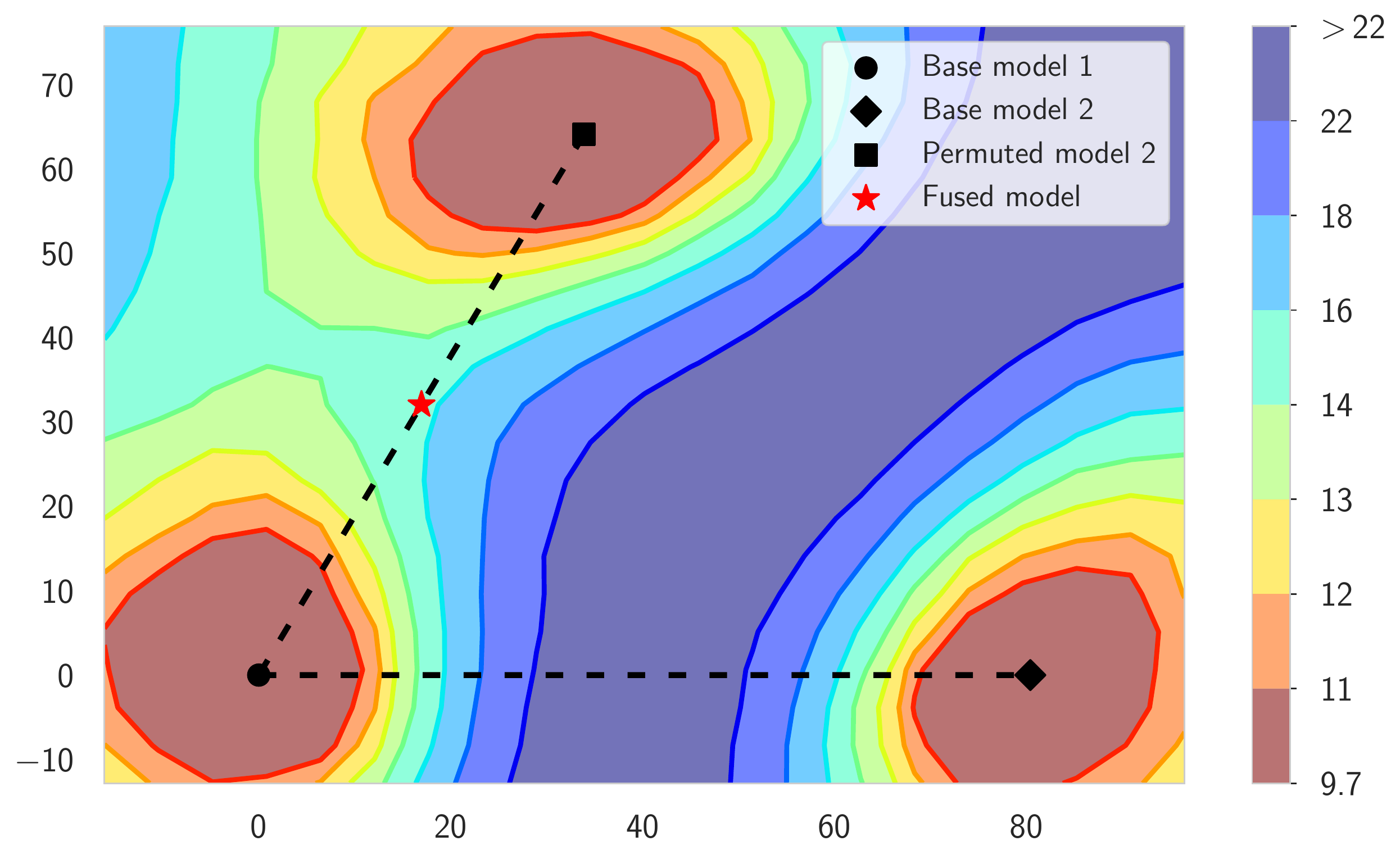}
    \end{subfigure}
\caption{The test error surface of (\textbf{Left}) RNN trained on SST-2 dataset, (\textbf{Right}) LSTM trained on SST-2 dataset.}
\label{fig: visualization for SST dataset}
\end{figure}

\end{document}